\documentclass[10pt,twocolumn,letterpaper]{article}

\usepackage[pagenumbers]{cvpr} 

\usepackage{graphicx}
\usepackage{amsmath}
\usepackage{amssymb}
\usepackage{mathtools}
\usepackage{booktabs}
\usepackage{caption}
\usepackage{subcaption}
\usepackage{xcolor}

\definecolor{cvprblue}{rgb}{0.21,0.49,0.74}
\usepackage[pagebackref,breaklinks,colorlinks,citecolor=cvprblue]{hyperref}

\def\paperID{1529} 
\def\confName{CVPR}
\def\confYear{2024}

\title{Inpaint3D: 3D Scene Content Generation using 2D Inpainting Diffusion}

\author{
Kira Prabhu\textsuperscript{1*}\hspace{0.3mm}
Jane Wu\textsuperscript{2*\dag}\hspace{0.3mm}
Lynn Tsai\textsuperscript{1*}\hspace{0.3mm} 
Peter Hedman\textsuperscript{1}\hspace{0.3mm}
Dan B Goldman\textsuperscript{1}\hspace{0.3mm}
Ben Poole\textsuperscript{1}\hspace{0.3mm}
Michael Broxton\textsuperscript{1}\\
\textsuperscript{1}Google Research \quad \textsuperscript{2}Stanford University \\
}

\begin{document}

\twocolumn[{%
\renewcommand\twocolumn[1][]{#1}%
\maketitle
\setcounter{figure}{0}
\begin{center}
    \centering
    \captionsetup{type=figure}
    \includegraphics[width=0.95\linewidth]{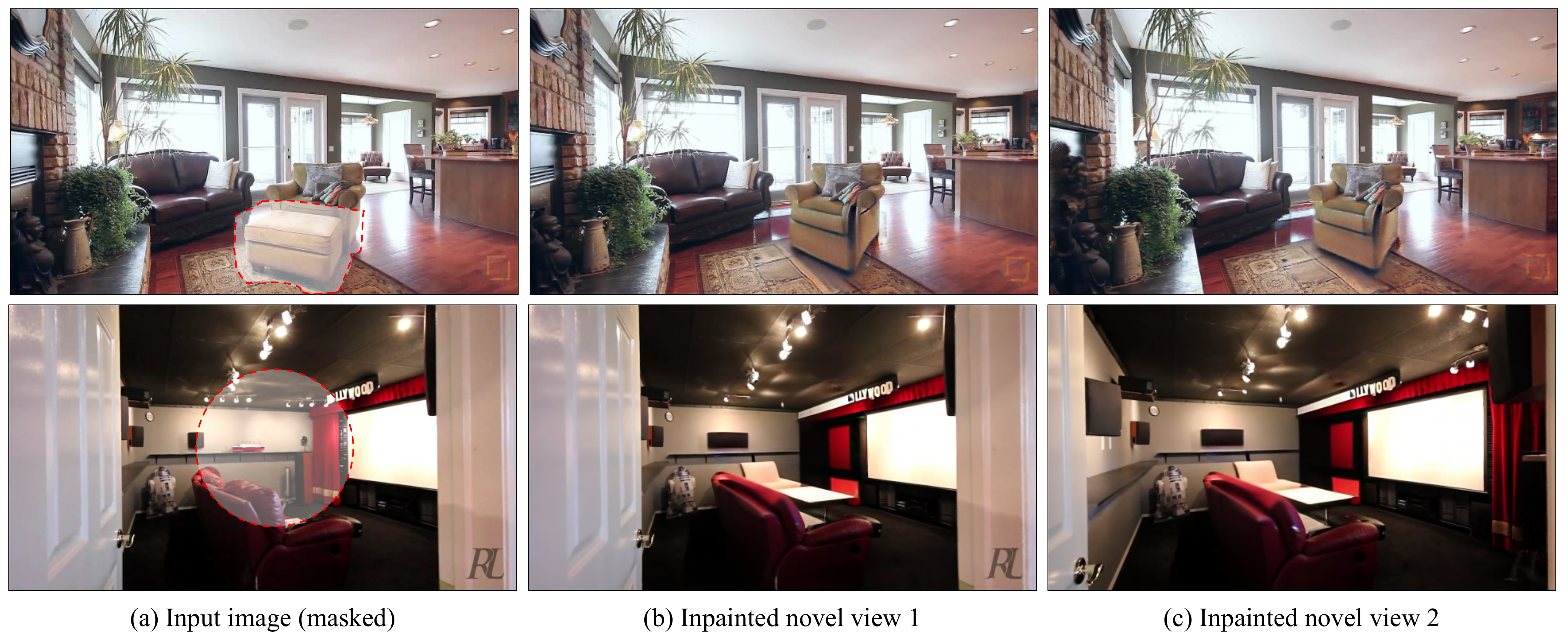}
    \captionof{figure}{Our method generates realistic scene content in arbitrary masked 3D regions (a) by distilling priors from a 2D inpainting diffusion model. Rendering the optimized 3D representation (a NeRF) from novel viewpoints (b, c) highlights the semantic and geometric consistency of the inpainted content with the rest of the  scene. }\label{fig:opening_figure_v2}
\end{center}
}]

\let\thefootnote\relax\footnotetext{\textsuperscript{*} Equal contribution.}
\let\thefootnote\relax\footnotetext{\textsuperscript{\dag}
Work done at Google.}

\begin{abstract}
This paper presents a novel approach to inpainting 3D regions of a scene, given masked multi-view images, by distilling a 2D diffusion model into a learned 3D scene representation (e.g.\ a NeRF). Unlike 3D generative methods that explicitly condition the diffusion model on camera pose or multi-view information, our diffusion model is conditioned only on a single masked 2D image. Nevertheless, we show that this 2D diffusion model can still serve as a generative prior in a 3D multi-view reconstruction problem where we optimize a NeRF using a combination of score distillation sampling and NeRF reconstruction losses. Predicted depth is used as additional supervision to encourage accurate geometry. We compare our approach to 3D inpainting methods that focus on object removal. Because our method can generate content to fill any 3D masked region, we additionally demonstrate 3D object completion, 3D object replacement, and 3D scene completion.  
\end{abstract}

\section{Introduction}
Diffusion models have seen widespread and rapid adoption thanks to their ability to synthesize remarkably high quality 2D images \cite{sohl2015deep,ho2020denoising,kingma2021variational,ho2022cascaded,ho2022imagen,song2020score,dhariwal2021diffusion,ho2022cascaded,rombach2022high,saharia2022image,saharia2022photorealistic} and videos \cite{ho2022imagen,ho2022video}. One  popular application of these models is 2D inpainting and outpainting, which generates content in a user-selected region~\cite{yu2019free,saharia2022palette}. Because this process is simple, powerful, and often seamless, generative inpainting has become a fundamental tool in many open source and commercial image generation workflows \cite{park2019semantic}.

While the toolset for 2D generative imaging is maturing rapidly, the extension of diffusion-based 2D generative methods to 3D content creation workflows is still a newly emerging field. Three-dimensional content is significantly more challenging to generate because the result must be consistent and semantically correct across a range of 2D image viewpoints. Furthermore, training diffusion models to generate 3D scenes directly via the diffusion process is generally impractical due to the lack of suitably diverse and plentiful 3D training data. Instead, methods like score distillation sampling (SDS) \cite{poole2022dreamfusion} and other variations \cite{wang2023prolificdreamer,katzir2023noise,meng2023distillation,warburg2023nerfbusters} have made progress in this space by using existing 2D diffusion models as a generative prior for 3D synthesis. This is accomplished by optimizing a 3D representation like a Neural Radiance Field (NeRF) using the constraint that 2D viewpoints rendered from the NeRF must be well explained by the conditional probability distribution modeled by the diffusion model. The distillation loss has a straightforward formulation via the probabilistic score (i.e. the gradient of the log-likelihood function).

Distillation sampling offers a way to leverage 2D diffusion models and bridge the gap from 2D to 3D, but it has two main drawbacks that limit its flexibility. First, SDS can be unstable without additional rendering cues such as a lighting model \cite{poole2022dreamfusion}. Subsequent work has shown that explicitly conditioning the diffusion model on 3D cues like multi-view data \cite{yoo2023dreamsparse} or camera pose information \cite{gu2023nerfdiff,liu2023zero,zhou2023sparsefusion, watson2022novel} can be critical to obtaining high quality results. Explicit 3D conditioning necessitates training a custom 3D-aware diffusion model, which requires large quantities of multi-view, calibrated image data.
A second limitation is that the conditioning signal the diffusion model was trained with must be well-defined during SDS optimization, which is not straightforward to formulate in the case of 3D inpainting.
Rather than generating 3D content directly, recent work in 3D generative inpainting takes the approach of inpainting individual 2D images first before fusing the inpainted images together in a 3D representation (as a NeRF) \cite{mirzaei2023spin, mirzaei2023reference, wang2023inpaintnerf360}.
Such approaches are limited by the quality of the pretrained 2D inpainting model, whereas SDS is formulated as an optimization problem that enables greater flexibility (by continuously sampling the diffusion prior model).

In this paper we seek to address the drawbacks discussed above by combining score distillation sampling (SDS) \cite{poole2022dreamfusion} with traditional NeRF 3D reconstruction losses into a joint optimization framework. We mask the SDS loss to target only the region to be inpainted, while the reconstruction losses target the remaining known regions visible and unmasked in the input images. The mask can be flexibly defined either in the reconstruction space (as a bounding volume) or in image space (as a multi-view mask). Unlike other approaches that rely on text-conditioned diffusion models, we train a 2D inpainting diffusion model whose only conditioning input is a masked 2D image. In particular, we note that our diffusion model does not make use of any explicit 3D conditioning or rendering cues as in prior work.
We trained our diffusion model on the RealEstate10k Dataset \cite{zhou2018stereo}, which consists of a large corpus of motion sequences capturing static indoor scenes. By tailoring the pre-trained diffusion model to the downstream tasks and datasets of interest, i.e.\ via inpainting, we are able to define a score distillation loss that encourages synthesized content that resembles samples from the diffusion model in the masked region. Experiments demonstrate that our method is able to generate semantically and geometrically consistent 3D content given initial views of a scene, and we compare these results to a state-of-the-art 3D inpainting approach \cite{mirzaei2023spin}.

\section{Related Work}\label{sec:related_work}
Inpainting is defined as filling masked regions of an image with realistic content \cite{bertalmio2000image}. Generative models are the current state-of-the-art in 2D inpainting \cite{yu2019free,saharia2022palette}, as diffusion model samples are diverse and they excel at matching the image context surrounding the inpainting mask. However, inpainting in 3D is far more challenging. There are some non-learned methods that leverage multi-view consistency, patch-matching, and various heuristics to inpaint in 3D~\cite{thonat2016multi,Baek_2016_CVPR,le2018light,philip2018plane}, but these lack robustness and fall far short of the quality that is now possible with 2D inpainting using diffusion models. Researchers have recently discovered several approaches that bring diffusion-based generative imaging into three dimensional problems. In this section we will survey these methods in the context of consistent 3D inpainting.

\begin{figure*}[t]
    \centering
    \includegraphics[width=\linewidth]{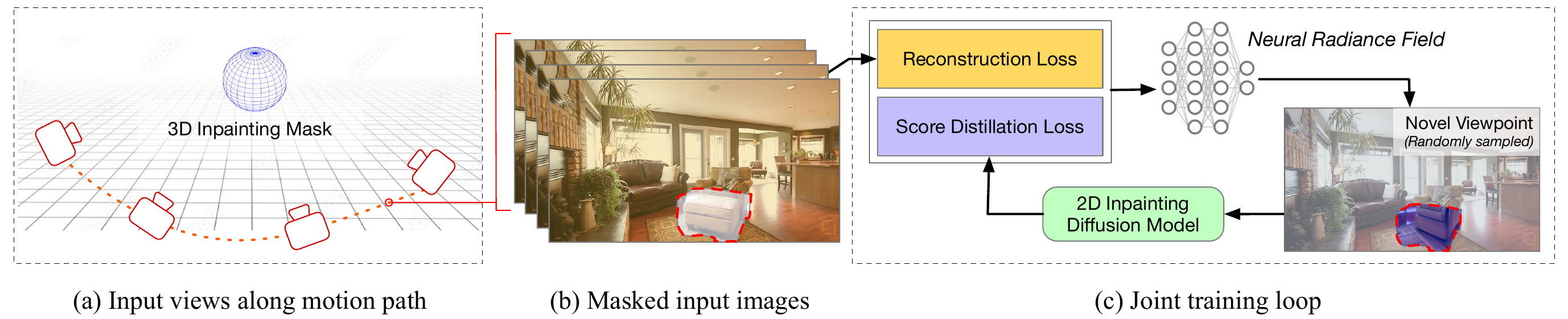}
    \caption{Overview of the proposed method. Given input images of a static scene and a 3D inpainting mask, the images are masked and a NeRF is optimized using our joint synthesis and reconstruction optimization scheme. In particular, the unmasked regions of the scene are supervised via a color reconstruction loss, and the masked regions are constrained using score distillation sampling with a frozen image inpainting model (see Figure \ref{fig:sds} for more details). The optimized NeRF can be rendered to novel views along with predicted depth maps.}
    \label{fig:overview}
\end{figure*}

\noindent \textbf{Native 3D diffusion:} A straightforward approach to 3D inpainting would be to generate volumetric content directly using a diffusion model. If this were possible, then methods for inpainting with 2D diffusion models would translate directly into 3D problems. However, diffusion models must be trained with large datasets that contain numerous diverse examples of the data distribution they are modeling. It can be challenging to find enough data to train a 3D diffusion model; widely used real world 3D datasets such as ShapeNet \cite{saharia2022palette} and Omni3D \cite{brazil2023omni3d} (or simulation data) can be used \cite{muller2023diffrf,shue20233d,warburg2023nerfbusters} to model highly object-centric 3D distributions, but few datasets are available for modeling 3D real world scenes. Training a 3D diffusion model also requires careful consideration of how Gaussian noise is applied as part of the diffusion Markov process. A number of recent works \cite{vahdat2022lion,wang2023rodin,anciukevivcius2023renderdiffusion} train latent diffusion models such that noise is applied in a lower-dimensional latent space, circumventing the problem of how to apply noise to the chosen 3D parameterization, but this too has only been explored for generating isolated objects or people.

\noindent \textbf{Independent 2D diffusion and 3D reconstruction:} Some approaches sidestep the difficulties of working with 3D diffusion models by independently inpainting the individual images in a multi-view dataset using 2D diffusion models, and then aggregating these into a 3D representation. The obvious problem encountered by these methods is that inpainting results have no guarantee of 3D consistency in the different images, and this must be reconciled using depth maps \cite{wei2023clutter,mirzaei2023spin, mirzaei2023reference, liu2022nerfin}, estimated normals~\cite{kasahara2023ric}, confidence metrics \cite{weder2023removing}, robust perceptual losses~\cite{mirzaei2023spin,wang2023inpaintnerf360} or -- for the adjacent task of 3D scene editing -- by interleaving diffusion with 3D reconstruction~\cite{instructnerf2023}. These methods focus on object removal tasks where the scene does not have significant depth complexity and the background is well observed enough that inpainting results are similar across the multi-view dataset. In contrast, our work focuses on general 3D inpainting, where there is no straightforward separation into foreground and background regions, making it necessary to complete complex scene geometry in a plausible 3D consistent way.

\noindent \textbf{2D to 3D Distillation:} To leverage the strong generative priors learned by 2D diffusion models in 3D, a number of methods have been proposed for distilling diffusion priors to downstream optimization tasks \cite{poole2022dreamfusion,jain2022zero,katzir2023noise,wang2023prolificdreamer, wang2023score, Lin_2023_CVPR}. But these models rely on text-conditioned diffusion models, and are therefore awkward to adapt for inpainting tasks. Also, achieving high quality reconstructions using these methods often require conditioning or otherwise providing the diffusion model with some 3D cues, such as camera pose information~\cite{tseng2023consistent}, multi-view conditioning~\cite{zhou2023sparsefusion}, a 3D feature volume~\cite{chan2023generative}, monocular depth prediction~\cite{fridman2023scenescape, xiang20233d}, or inferred or explicit geometry priors~\cite{yoo2023dreamsparse, Metzer_2023_CVPR}. The joint optimization framework we propose in this paper builds on this distillation approach, and specifically on score distillation sampling~\cite{poole2022dreamfusion}. But rather than condition on text or explicit 3D queues as described in the works above, we condition only on a 2D masked image. This immediately unlocks the ability to perform inpainting (by projecting a 3D mask into the 2D image), while our joint training framework ensures 3D consistency. In the next section we turn to a detailed description of this joint optimization framework.

\section{Background}\label{sec:background}
Section \ref{sec:nerf} provides background on the NeRF model that we use to parameterize 3D space, and Section \ref{sec:2d_priors} provides preliminaries for diffusion models and score distillation sampling (SDS). Section \ref{sec:joint_training} describes how we combine these techniques to formulate our novel 3D inpainting approach.

\begin{figure*}[ht]
    \centering
    \includegraphics[width=\linewidth]{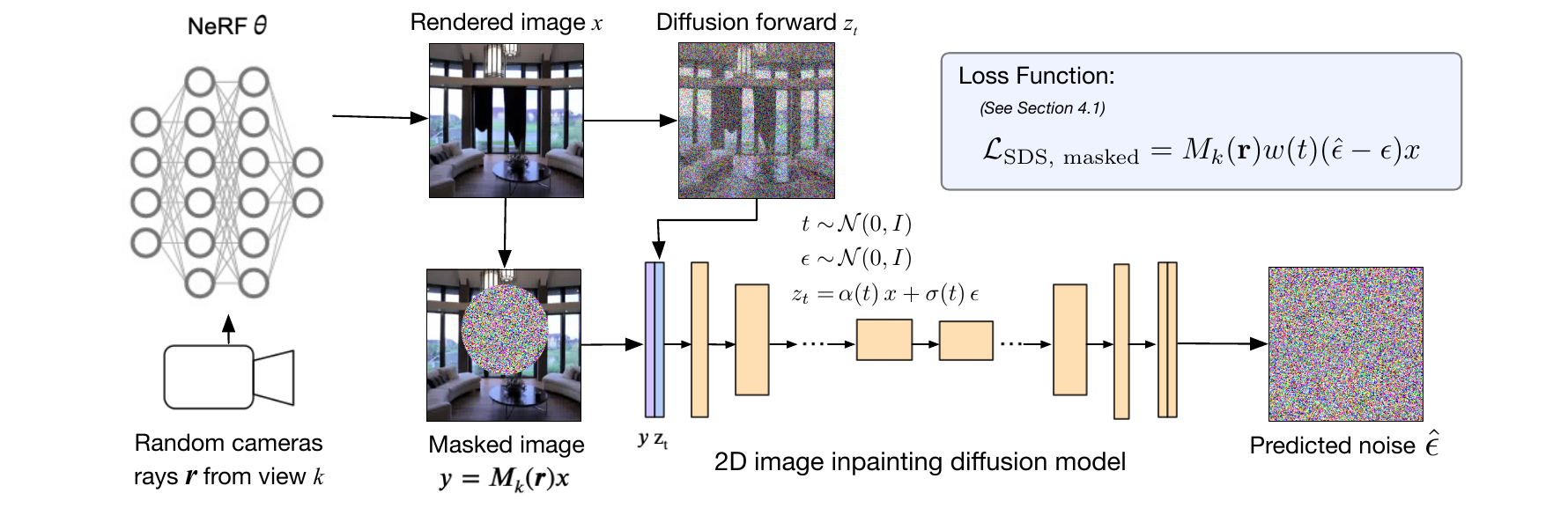}
    \caption{One step of our score distillation sampling approach applied to masked regions.}
    \label{fig:sds}
\end{figure*}

\subsection{3D Representation}\label{sec:nerf}
The geometry and appearance of each scene is modeled as a NeRF \cite{mildenhall2021nerf} parameterized by a multilayer perceptron (MLP).
A NeRF is a learned 3D representation that maps any location in world space to RGB color $\vec{c}$ and density $\sigma$.
To render a NeRF from a given camera to obtain an RGB image, (differentiable) volumetric raytracing is typically used.
For each pixel, a ray defined by $r(t) = r_0 + tr_d$ is cast from the camera aperture $r_0$ through the pixel center and 3D points are sampled along each ray.
These points $v_i$, where $i\in \{1, \ldots,  N\}$, are passed to the MLP to obtain output colors $\vec{c}_i$ and densities $\sigma_i$.
The colors and densities are alpha-composited to obtain the final rendered RGB image with pixels $\Tilde{I}(r)$:
\begin{equation}
    \Tilde{I}(r) = \sum_{i=1}^N w_i(1-\exp(-\sigma_i\delta_i))\vec{c}_i
\end{equation}
\begin{equation}
    w_i = \exp\left(- \sum_{j=1}^{i-1}\sigma_j\delta_j\right)\quad\quad \delta_i = ||v_{i+1} - v_i ||_2
\end{equation}
Similarly, a corresponding depth map $\Tilde{D}(r)$ can be computed by accumulating density along each ray \cite{niemeyer2022regnerf}:
\begin{equation}\label{eq:nerf_depth}
    \Tilde{D}(r) = \sum_{i=1}^N w_i(1-\exp(-\sigma_i\delta_i))t_i
\end{equation}
where $t_i = (v_i - r_0)/r_d$.

NeRFs are typically trained by optimizing MLP parameters per-scene given a set of input images $I_k$ and corresponding camera parameters. 
The NeRF implementation we use is from Zip-NeRF \cite{barron2023zip}, which builds on improvements made by MipNerf360 \cite{barron2021mip} and Instant-NGP \cite{muller2022instant} to provide anti-aliasing and faster training than the original NeRF implementation.

\subsection{Diffusion Models}\label{sec:2d_priors}
Diffusion models \cite{sohl2015deep, song2020score, ho2020denoising} learn a latent-variable generative model through a sequential generation process. Starting with the data distribution $z_0 \sim q(z_0)$, a sequence of latent variables $z_1, \ldots, z_T$ can be constructed through a forward diffusion process $q(z_{1:T}|z_0)$. The forward process adds increasing amounts of Gaussian noise, where each latent $z_t$ can be constructed from a datapoint $z_0$ and random noise $\epsilon \sim\mathcal{N}(0, I)$:
 \begin{equation}
    z_t(z_0, \epsilon) = \alpha(t)z_0 + \sigma(t)\epsilon,
    \label{eq:diffusion_fwd}
 \end{equation}
where $\alpha(t)$ and $\sigma^2(t)$ are coefficients chosen to reduce the signal-to-noise ratio of the data as $t$ increases while preserving the variance. 
A diffusion model is trained to reverse this process and thus learn a generative model 
that synthesizes samples from noise. 
Following \cite{ho2020denoising}, learning the reverse process reduces to learning a function to predict the noise that was added to the data $\hat{\epsilon}_\phi$:
\begin{equation}
    \mathcal{L}_{\text{diffusion}} = \mathbb{E}_{\epsilon,t}\left[||\hat{\epsilon}_\phi\left(\alpha(t)z_0 + \sigma(t)\epsilon; t\right) - \epsilon||^2\right].
\end{equation}

We trained a 2D inpainting diffusion model to predict $\hat{\epsilon}_{\phi}(z_t; t, y)$, where $y$ is a masked image.
The model was trained using the RealEstate10k dataset \cite{zhou2018stereo}, which contains videos of static indoor and outdoor scenes.
See Section \ref{sec:diffusion_model_training} and the appendix for more implementation details.

\subsection{Score Distillation Sampling}\label{sec:sds}
Score distillation sampling (SDS) \cite{poole2022dreamfusion} uses a pretrained diffusion model as a prior for optimizing parameters $\theta$ of a generator function $x=g(\theta)$ that outputs samples resembling the learned data distribution of the diffusion model.
The SDS loss gradients w.r.t.\ $\theta$ are defined as
\begin{equation}\label{eq:sds_loss}
    \nabla_{\theta}\mathcal{L}_{\text{SDS}} \coloneqq \mathbb{E}_{t,\epsilon} \left[w(t)(\hat{\epsilon}_{\phi}(z_t;y,t) - \epsilon)\frac{\partial x}{\partial \theta}\right]
\end{equation}
where in practice $t$ is randomly sampled at each optimization step and $w(t)$ is some timestep-dependent weighting function.
In the context of our approach, a NeRF serves as our differentiable image generator, as in DreamFusion \cite{poole2022dreamfusion}. $x$ is a volume-rendered RGB image, $y$ is the result of masking $x$ with 3D inpainting mask via ray-mask intersection, $z_t$, is the output of a single forward diffusion step applied to $x$, and $\hat{\epsilon}_{\phi}(z_t;y,t)$ is the output of the pre-trained 2D inpainting diffusion model described in in Section \ref{sec:2d_priors}. 

\section{Method}\label{sec:joint_training}
Given a set of multi-view images along a motion path $\mathcal{I} = \left\{ I_k \right\}_{k=1}^n$ and 3D-consistent image masks $\mathcal{M} = \left\{ M_k \right\}_{k=1}^n$, we aim to simultaneously synthesize realistic and 3D-consistent content in the masked regions of the reconstruction volume and accurately reconstruct the unmasked regions of the input images. Our joint SDS and reconstruction approach expands upon the foundational ideas in Section \ref{sec:background}. See Figure \ref{fig:overview}.

It is not immediately evident how to use SDS to distill priors from a 2D inpainting diffusion model to guide 3D synthesis.
In DreamFusion \cite{poole2022dreamfusion}, for instance, a text prompt is taken as input and provides a fixed conditioning signal to the diffusion model during NeRF optimization.
By contrast, inpainting diffusion models introduce a constraint that the conditioning image $y$ should be a masked version of the data sample $x$ that is passed through the forward diffusion process (and inpainted).
Thus, since we are inpainting the 3D scene and not the 2D input images, we use the \textit{rendered} images as the conditioning signal.

As discussed in Section \ref{sec:nerf}, we choose a NeRF as our 3D scene representation and differentiable image generator for SDS.
The unmasked regions of the 3D scene are supervised using the RGB values from the input images, which ensures faithfulness to the ground truth scene while maintaining consistent geometry. 
Priors from our pre-trained 2D inpainting diffusion model (see Section \ref{sec:2d_priors}) are distilled via SDS to inpaint new content in the masked region. The joint training step is illustrated in column (c) of Figure \ref{fig:overview}. Implementation details pertaining to our diffusion model and optimization parameters are provided in Section \ref{sec:diffusion_model_training}.

\subsection{Joint Synthesis and Reconstruction}
We assume that the input images $\mathcal{I}$ have known camera parameters and define a mask function $M_k(\mathbf{r})$ that returns one for each ray in a set $\mathbf{r}$ that intersects with the masked region and zero otherwise.
To reconstruct the unmasked regions of the scene, we sample random rays $\mathbf{r}_k$ across all input images at each optimization step to obtain estimated colors $\Tilde{I}_k(\mathbf{r}_k)$ from the NeRF.
A NeRF loss is then applied only to rays intersecting unmasked pixels:
\begin{equation}\label{eq:nerf_unmasked}
\small
    \mathcal{L}_{\text{recon, unmasked}} = \sum_k \left(1-M_k(\mathbf{r}_k)\right)\rho(\Tilde{I}_k(\mathbf{r}_k) - I_k(\mathbf{r}_k))
\end{equation}
where $\rho(x, x^*) = \sqrt{(x - x^*)^2 -\epsilon^2}$ and $\epsilon=0.001$ (the Charbonnier loss \cite{charbonnier1994,barron2021mip}).

In the masked regions, we apply SDS as follows (see Figure \ref{fig:sds}). First, we sample $256\times 256$ patches from the rendered images $\Tilde{I}_k$ that contain some overlap with the inpainting masks. Each rendered patch $x_k$ is masked according to $M_k(r_k)$ in order to obtain the diffusion conditioning signal $y_k=M_k(r_k)x_k$. Per-view diffusion timestep $t_k$ and noise $\epsilon_k$ are randomly sampled, and $z_{k,t}$ is computed via 
$z_{k,t} = \alpha(t_k)x_k + \sigma(t_k)\epsilon_k$ (as in Equation \ref{eq:diffusion_fwd}).
The diffusion model inputs $z_t, y, t_k$ are passed to the pre-trained inpainting diffusion model to obtain $\hat{\epsilon}_{\phi}(z_t;y,t_k)$, and the SDS loss as defined in Equation \ref{eq:sds_loss} is applied via
\begin{equation}
\small
    \mathcal{L}_{\text{SDS, masked}} = w(t)\sum_k M_k(r_k)(\hat{\epsilon}_{\phi}(z_{k,t};y_k,t_k) - \epsilon_k) x_k
\end{equation}

Finally, we apply additional volume rendering losses to both masked and unmasked rays. These include distortion and interlevel losses from Mip-Nerf 360, as well as the patch-based depth smoothness loss, $\mathcal{L}_{\text{DS}}$, from RegNeRF \cite{niemeyer2022regnerf} (modified to include a bilateral weighing term guided by the predicted colors). Thus, the total loss function is
\begin{equation}
\begin{split}
    \mathcal{L}_{\text{total}} &= w_1\mathcal{L}_{\text{recon, unmasked}} +  w_2\mathcal{L}_{\text{SDS, masked}} \\
    &\quad + w_3\mathcal{L}_{\text{dist}} + w_4\mathcal{L}_{\text{inter}} + w_5\mathcal{L}_{\text{DS}}
\end{split}
\end{equation}
where $w_{1:5}$ are loss weights.
In practice, we found that $w_2$ needs to be small relative to the other loss weights.

\subsection{Optimization Details}
At the beginning of optimization, only Equation \ref{eq:nerf_unmasked} is computed to provide a more stable initialization of the MLP for applying SDS. We also fix the input view $k$ within a small baseline selected for SDS for a number of steps in order to converge on the inpainted content before handling multi-view consistency. Afterwards, the input views are randomly selected.

To encourage 3D consistency while still allowing for view-dependent effects, we decompose the RGB NeRF prediction into diffuse and sparse components for the final RGB color prediction, where the diffuse component is predicted by a small MLP taking only the features computed from the input 3D point (but not ray direction). At the start of optimization, we optionally use the diffuse component of the RGB image as input to the diffusion model to encourage consistent SDS updates across views, before annealing to both diffuse and specular components. We empirically observed this annealing increased the level of detail in the synthesized images for RealEstate10k scenes, where highly reflective materials are common.

In order to encourage sample diversity towards the beginning of optimization and generate consistent diffusion model outputs towards the end, we anneal the diffusion timestep such that at any optimization step $s \in [0, S]$, the value of $t$ is sampled from 
\begin{align}
    p(t) &\sim \mathcal{U}(\eta - t_{\text{inv}}, \eta) ~~ \text{, where} \\
    \eta &= \max(t_{\text{min}}, t_{\text{max}}(1 - s/S))
\end{align}
We use $t_{\text{min}}=0.2$, $t_{\text{max}}=0.95$, and $t_{\text{inv}}=0.2$ for all experiments. We optimize the NeRF over 16 GPUs, where at each step, each GPU takes an optimization step in parallel.

\section{Experiments}
Our implementation uses a variant of Mip-Nerf 360, backed by Instant-NGP hash grids, as our view-synthesis backend. Specifically, we use the scene contraction and proposal MLPs from Mip-Nerf 360, hash grids with coarsest and finest resolutions of 16 and 8192, and model exposure and scene lighting variations using GLO codes from NeRF-W \cite{martinbrualla2020nerfw}.

\begin{figure*}[ht]
\includegraphics[width=\linewidth]{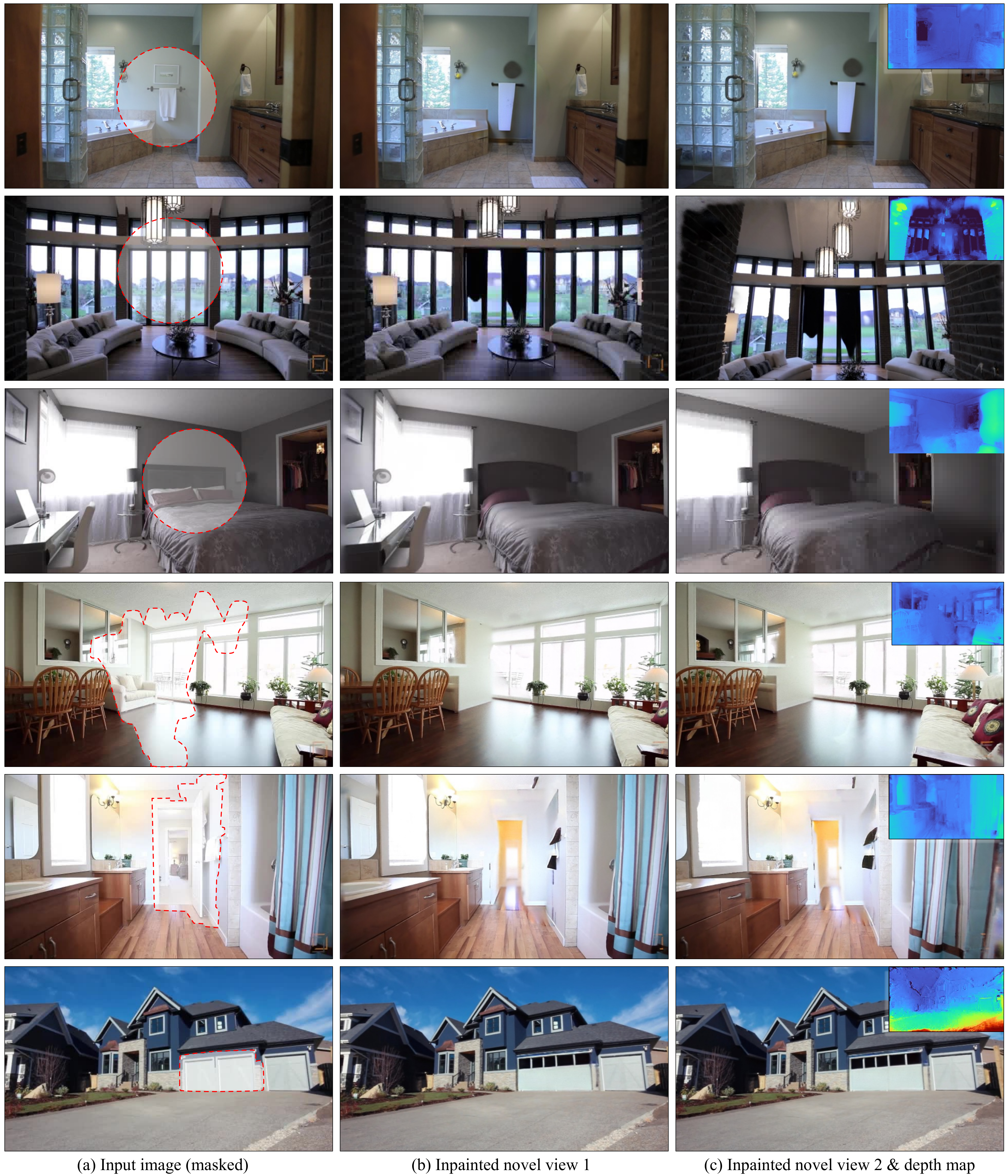}
    \caption{3D reconstructions of inpainted RealEstate10k scenes and NeRF depth maps rendered from novel views along the training path (b, c). The 3D inpainting mask is highlighted in (a). Our inpainting method is able to synthesize realistic content in masked regions varying in size, shape, location, and occlusions.}
    \label{fig:inpainting_results}
\end{figure*}

\begin{figure}[ht]
\centering
\includegraphics[width=\linewidth]{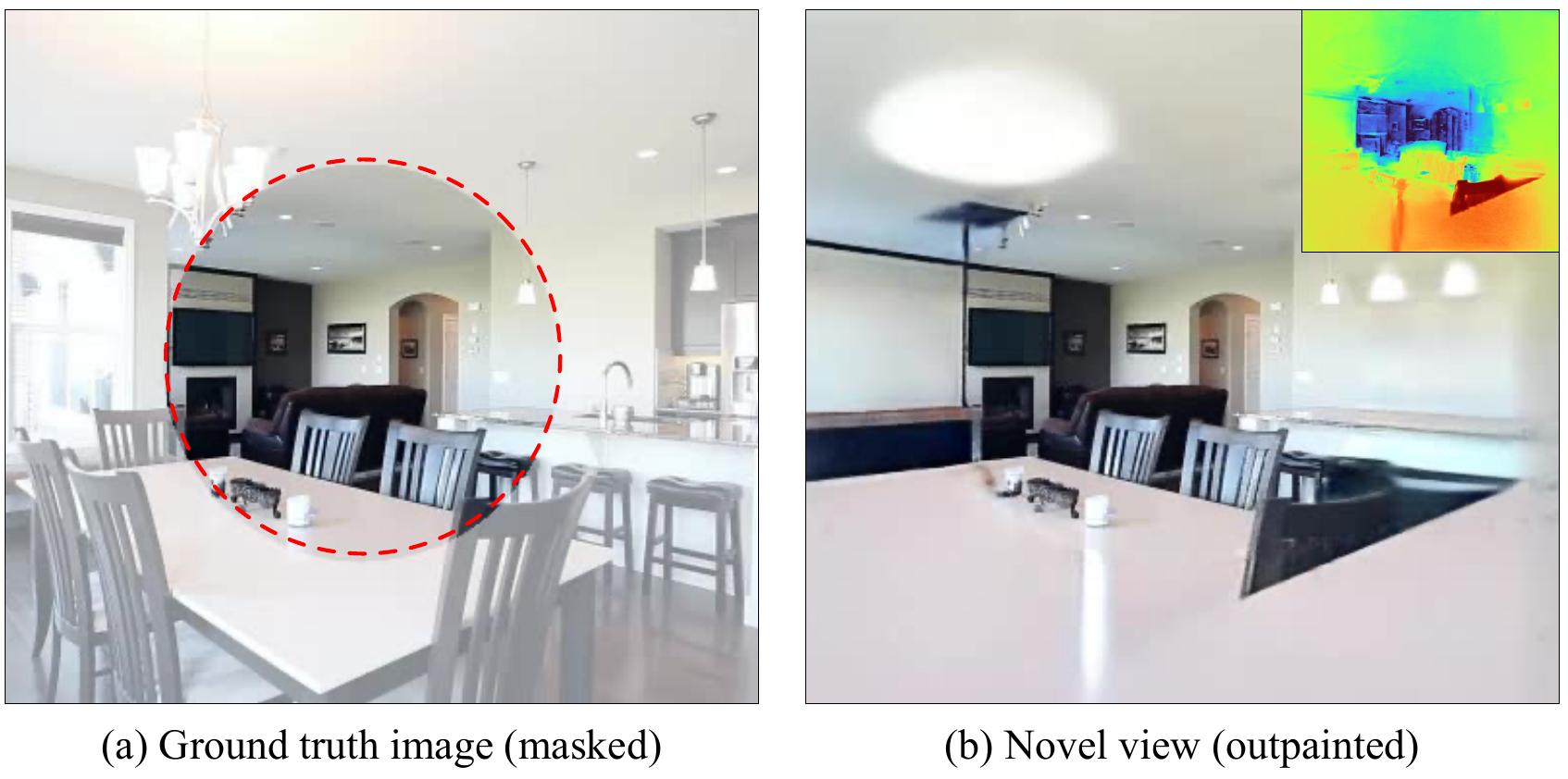}
\caption{Outpainting can be performed by inverting the mask. Our joint optimization will then generate novel scene content (b) and plausible depth (see inset) outside of the mask boundary.} \label{fig:outpainting}
\end{figure}

\subsection{Diffusion Model Training}\label{sec:diffusion_model_training}
We build on the network architecture from Palette \cite{saharia2022palette} for our inpainting diffusion model and use an input resolution of $256\times256$.
During training, each image is preprocessed by taking either a random 256px square crop or the largest center crop resized to 256px (similar to Palette \cite{saharia2022palette}), and a random inpainting mask is generated.
To generate inpainting masks, we follow \cite{yu2018generative,saharia2022palette} and use a combination of free-form strokes and rectangular masks.
The conditioning signal is the input image with the masked region filled with Gaussian noise (as in Palette \cite{saharia2022palette}).
Classifier-free guidance \cite{ho2022classifier} is used to enable tuning of sample quality vs. diversity for SDS.
The diffusion model is trained from scratch on the RealEstate10k dataset \cite{zhou2018stereo} (see Section \ref{sec:datasets} for more dataset details).

\subsection{Datasets}\label{sec:datasets}
We use a subset of the RealEstate10k dataset \cite{zhou2018stereo} for training the inpainting diffusion model and use a separate held-out test set for evaluating our 3D inpainting method.
In total, the dataset consists of 10 million frames from 10,000 YouTube videos, and includes static indoor and outdoor scenes.
The native resolution of the images is $720\times1280$. The images were resized to $1080\times1920$ and random $256\times256$ crops were used to train the diffusion model.
For 3D inpainting with our joint synthesis and reconstruction technique, the images were downsampled to $360\times640$.

To compare with existing work in object removal (a subset of 3D inpainting scenarios), we also evaluate on the SPIn-NeRF dataset \cite{mirzaei2023spin}, which consists of 10 static outdoor scenes captured from 60 views.
Each frame also has a corresponding object mask that was manually annotated.
Additionally, 40 views of each scene without the object were captured for evaluation.

\subsection{3D Inpainting Masks}\label{sec:inpainting_masks}
We demonstrate the robustness of our approach to different 3D inpainting tasks by generating a variety of masks.

\noindent \textbf{Sphere Masks:} We position a sphere at a fixed distance along the optical axis of the center camera in the input path. We compute sphere-ray intersections at each input viewpoint in order to determine the projection of the sphere onto each input image. With these masks, we illustrate that we can inpaint sizable 3D volumes of a scene with new content that is both 3d-consistent and makes semantic sense.

\noindent \textbf{Object masks:} We first pre-train a NeRF in order to obtain depth information about the scene. For a single view, we then select points which identify the objects we wish to mask, and provide these as input to the Segment-Anything (SAM) model \cite{kirillov2023segment}. We choose the mask with the highest SAM score and reproject it to the other input views using the estimated NeRF depths. Finally, we dilate the masks in order to fill holes and smooth edges. These masks demonstrate the ability of our approach to be used for tasks such as object removal and replacement.

\noindent \textbf{Scribble masks:} We draw a random path through an input image, and dilate the points along the path with ellipses. As with the object masks, we first generate the mask in a single input view, then use the estimated depth information from a pre-trained NeRF to project it to other input views. These masks present an interesting set of challenges in that they often span a variety of depths from background to foreground, and partially occlude objects.

\noindent \textbf{Outpainting masks:} These masks are created by inverting the sphere masks, such that we maintain only a sphere in the center of the image as scene context. The inpainting model then outpaints the majority of the scene. See Figure \ref{fig:outpainting}.

\subsection{Qualitative Results}
\begin{figure}[tb]
\centering
\includegraphics[width=\linewidth]{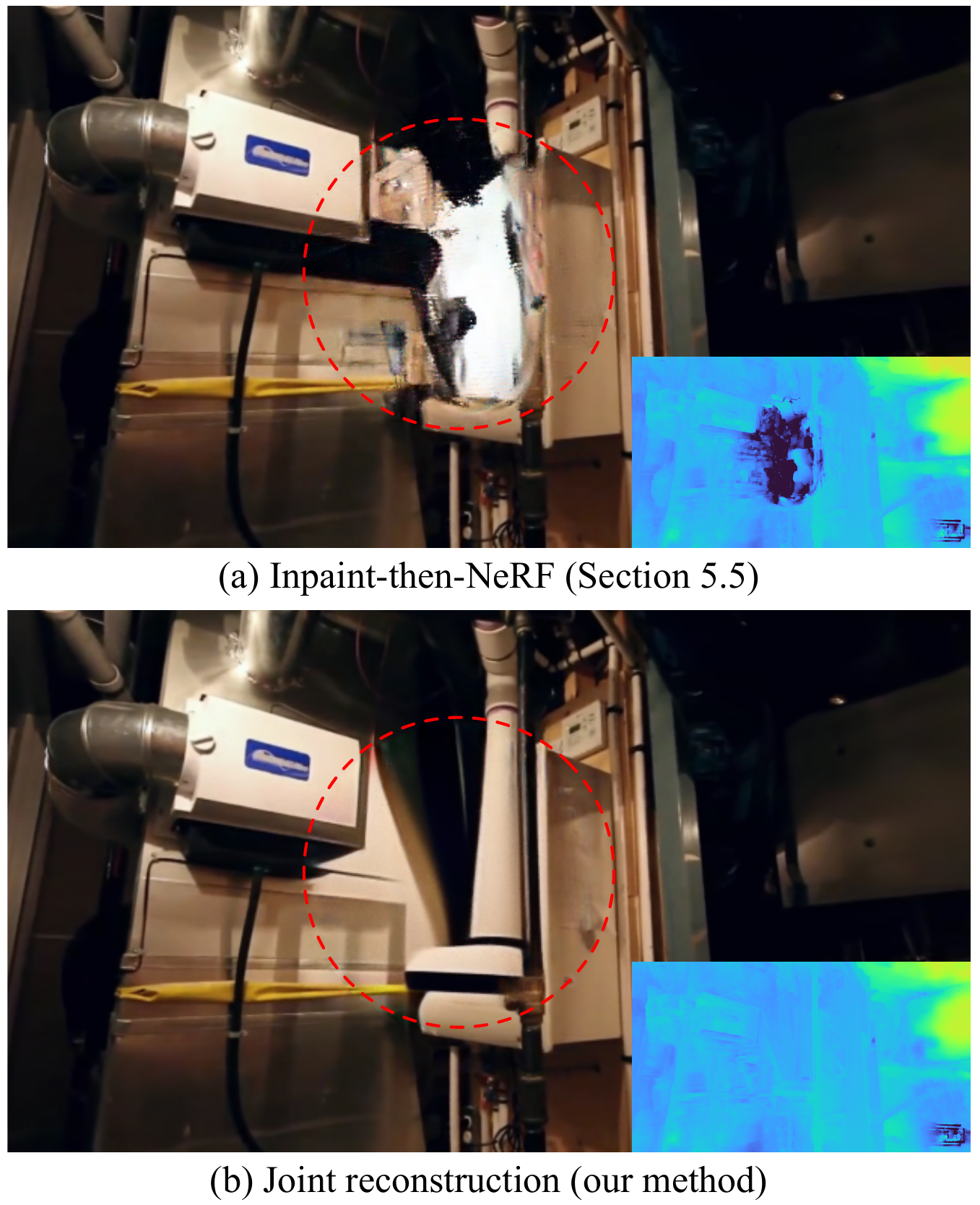}
\caption{Inpaint-then-NeRF Experiment. (a) Using our RealEstate10k diffusion model, we inpainted a masked region of our source 2D images from the motion path \textit{before} reconstructing a NeRF. (b) This led to lower quality inpainting and significantly worse multi-view consistency compared to joint reconstruction.}
\label{fig:ancestral_sampling_experiment}
\end{figure}

\begin{figure*}[ht]
    \centering
    \includegraphics[width=0.8\linewidth]{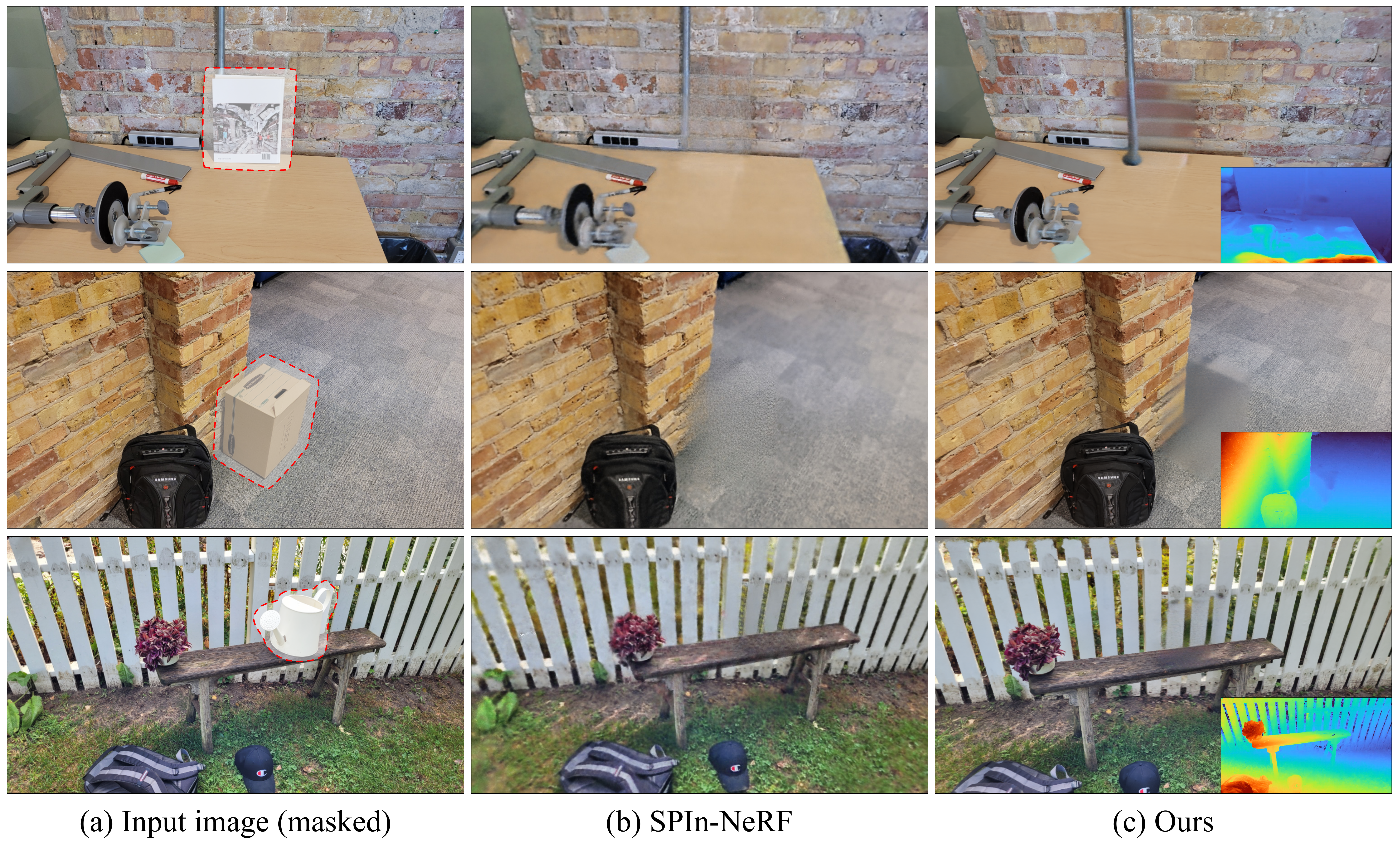}
    \caption{Qualitative comparison of our 3D inpainting method SPIn-NeRF where all images are rendered to a novel view (input images with the object included are shown in (a)).}
    \label{fig:inpainting_comparisons}
\end{figure*}
In Figure \ref{fig:inpainting_results}, we evaluate our 3D inpainting approach on examples from the RealEstate10k dataset. For each video, 80 consecutive frames are selected and a 3D inpainting mask is applied to each image. Column (a) shows the masked ground truth image corresponding to the first training view, and (b) and (c) show the optimized NeRF color and depth rendered from the different views along the training path.

We show masks which partially occlude objects in the scene (rows 1 and 2), and we also fully mask single objects in the background (row 3) and the foreground (row 4).
In these masked regions, we observe that our method is able to complete partially occluded furniture (e.g.\ tables, chairs) and  preserve visual properties of the scene (e.g.\ lighting, reflections). The results demonstrate that our method is robust over a variety of mask sizes, shapes, locations, and occlusions while also correctly reconstructing the unmasked regions of the input images in accordance with existing NeRF literature \cite{barron2023zip}.

We found that lower classifier-free guidance \cite{ho2022classifier} and SDS loss weights were better suited to object removal and completion tasks (e.g.\ scribble masks), while higher weights encouraged more creative content generation appropriate for larger masked regions (e.g.\ sphere, object, outpainting masks). Thus these two parameters were tuned by mask type, but beyond this we perform no scene-specific tuning.

\subsection{Comparison with Inpaint-then-NeRF}
In this section we compare with the established strategy of first inpainting the input images independently, then using a robust perceptual loss to train a NeRF \cite{mirzaei2023spin,wang2023inpaintnerf360}. This experiment was similar to the independent 2D diffusion followed by 3D reconstruction methods discussed in Section~\ref{sec:related_work}, albeit without any mechanism ensuring multi-view consistency. As seen in Figure~\ref{fig:ancestral_sampling_experiment}, this strategy does create plausible images in isolation, but struggles to achieve 3D consistency within a large inpainting region.

Table \ref{tab:metrics} shows a quantitative comparison of our method with SPIn-NeRF \cite{mirzaei2023spin} on their data (described in Section \ref{sec:datasets}). The metrics are computed on a bounding box surrounding the masked region, so only the inpainted region is evaluated. Our advantage in the accuracy metrics SSIM and LPIPS, which have reference images, is a strong indication that our results exhibit 3D consistency in the inpainted regions. Figure \ref{fig:inpainting_comparisons} shows qualitative results of the comparison. The depth maps shown in column (c) reinforce the 3D consistency of our results. From column (b), we observe that SPIn-NeRF tends to replicate local texture (e.g.\ the floor texture in the middle row), while our method better creates scene structure in inpainted regions (e.g.\ wall is completed in the middle row). See the appendix for more examples.

\begin{table}[ht]
  \centering
  \begin{tabular}{@{}rcc@{}}
    \toprule
    Metric & SPIn-NeRF & Ours \\
    \midrule
    PSNR $\uparrow$ & 17.40 & 17.37 \\
    SSIM $\uparrow$ & 0.351 & 0.463 \\
    LPIPS $\downarrow$ & 0.540 & 0.515 \\
    FID $\downarrow$ & 194.67 & 226.04\\
  \end{tabular}
\caption{Quantitative evaluation of our approach with SPIn-NeRF~\cite{mirzaei2023spin}, using their dataset (see Section~\ref{sec:datasets})}
  \label{tab:metrics}
\end{table}

\section{Conclusion}
In this paper, we leverage 2D inpainting diffusion models for 3D generation without the need for multi-view data \cite{yoo2023dreamsparse} or camera poses \cite{gu2023nerfdiff,liu2023zero,zhou2023sparsefusion}.
We use score distillation sampling \cite{poole2022dreamfusion} combined with NeRF reconstruction losses in a novel joint optimization framework. By tailoring the pretrained diffusion model to the downstream tasks and datasets of interest, e.g.\ via inpainting, we are able to define a score distillation loss that encourages synthesized content that resembles samples from the diffusion model. This content fills the masked region while unmasked content is reconstructed normally.
We show that this approach works well across a variety of mask types for both inpainting and outpainting tasks. Critically, our diffusion model need not be conditioned on any explicit 3D information; we show successful distillation of 3D content from a 2D diffusion model conditioned only on a single 2D masked image. In making this design choice, we are able to go beyond the problem formulations of prior work \cite{mirzaei2023spin,wang2023inpaintnerf360,weder2023removing,warburg2023nerfbusters,wei2023clutter,yin2023or} and handle a wider range of inpainting settings.

\noindent\textbf{Limitations and Future Work:} 
For future work, we are interested in exploring the growing number of methods beyond score distillation sampling for distilling diffusion priors \cite{katzir2023noise,wang2023prolificdreamer}.
During score distillation sampling, randomness induced by the diffusion process often results in high variance in $\hat{\epsilon}(t)$ and thus guides the optimization towards ``mean''. As a result, generated content often lacks high frequency detail and the inpainted content is not particularly sensitive to different random seeds.
Additional 3D priors beyond using a NeRF representation could also help to further improve the level of detail in our inpainting results.

\section*{Acknowledgements}
The authors would like to thank Ryan Overbeck, Jason Lawrence, and Steve Seitz for their generous support of this project. We also thank Jascha Sohl-Dickstein for his participation in our project meetings and his very thoughtful comments and feedback on our approach. We are grateful to Ryan Overbeck and Ricardo Martin-Brualla for their feedback during the writing process, and to Clément Godard and Ben Mildenhall for pointers on how best to run Mip-Nerf 360 on various datasets. Finally, we thank Richard Tucker for his help with using the RealEstate10k dataset.
J. W. was supported in part by the Gerald J. Lieberman Graduate Fellowship, the NSF Mathematical Sciences Postdoctoral Fellowship, and the UC President's Postdoctoral Fellowship.

{\small
\bibliographystyle{ieeenat_fullname}
\bibliography{egbib}

\begin{thebibliography}{65}
\providecommand{\natexlab}[1]{#1}
\providecommand{\url}[1]{\texttt{#1}}
\expandafter\ifx\csname urlstyle\endcsname\relax
  \providecommand{\doi}[1]{doi: #1}\else
  \providecommand{\doi}{doi: \begingroup \urlstyle{rm}\Url}\fi

\bibitem[Anciukevi{\v{c}}ius et~al.(2023)Anciukevi{\v{c}}ius, Xu, Fisher,
  Henderson, Bilen, Mitra, and Guerrero]{anciukevivcius2023renderdiffusion}
Titas Anciukevi{\v{c}}ius, Zexiang Xu, Matthew Fisher, Paul Henderson, Hakan
  Bilen, Niloy~J Mitra, and Paul Guerrero.
\newblock Renderdiffusion: Image diffusion for 3d reconstruction, inpainting
  and generation.
\newblock In \emph{Proceedings of the IEEE/CVF Conference on Computer Vision
  and Pattern Recognition}, pages 12608--12618, 2023.

\bibitem[Baek et~al.(2016)Baek, Choi, and Kim]{Baek_2016_CVPR}
Seung-Hwan Baek, Inchang Choi, and Min~H. Kim.
\newblock Multiview image completion with space structure propagation.
\newblock In \emph{Proceedings of the IEEE Conference on Computer Vision and
  Pattern Recognition (CVPR)}, 2016.

\bibitem[Barron et~al.(2021)Barron, Mildenhall, Tancik, Hedman, Martin-Brualla,
  and Srinivasan]{barron2021mip}
Jonathan~T Barron, Ben Mildenhall, Matthew Tancik, Peter Hedman, Ricardo
  Martin-Brualla, and Pratul~P Srinivasan.
\newblock Mip-nerf: A multiscale representation for anti-aliasing neural
  radiance fields.
\newblock In \emph{Proceedings of the IEEE/CVF International Conference on
  Computer Vision}, pages 5855--5864, 2021.

\bibitem[Barron et~al.(2023)Barron, Mildenhall, Verbin, Srinivasan, and
  Hedman]{barron2023zip}
Jonathan~T Barron, Ben Mildenhall, Dor Verbin, Pratul~P Srinivasan, and Peter
  Hedman.
\newblock Zip-nerf: Anti-aliased grid-based neural radiance fields.
\newblock \emph{arXiv preprint arXiv:2304.06706}, 2023.

\bibitem[Bertalmio et~al.(2000)Bertalmio, Sapiro, Caselles, and
  Ballester]{bertalmio2000image}
Marcelo Bertalmio, Guillermo Sapiro, Vincent Caselles, and Coloma Ballester.
\newblock Image inpainting.
\newblock In \emph{Proceedings of the 27th annual conference on Computer
  graphics and interactive techniques}, pages 417--424, 2000.

\bibitem[Brazil et~al.(2023)Brazil, Kumar, Straub, Ravi, Johnson, and
  Gkioxari]{brazil2023omni3d}
Garrick Brazil, Abhinav Kumar, Julian Straub, Nikhila Ravi, Justin Johnson, and
  Georgia Gkioxari.
\newblock {Omni3D}: A large benchmark and model for {3D} object detection in
  the wild.
\newblock In \emph{CVPR}, Vancouver, Canada, 2023. IEEE.

\bibitem[Chan et~al.(2023)Chan, Nagano, Chan, Bergman, Park, Levy, Aittala,
  De~Mello, Karras, and Wetzstein]{chan2023generative}
Eric~R Chan, Koki Nagano, Matthew~A Chan, Alexander~W Bergman, Jeong~Joon Park,
  Axel Levy, Miika Aittala, Shalini De~Mello, Tero Karras, and Gordon
  Wetzstein.
\newblock Generative novel view synthesis with 3d-aware diffusion models.
\newblock \emph{arXiv preprint arXiv:2304.02602}, 2023.

\bibitem[Charbonnier et~al.(1994)Charbonnier, Blanc-Feraud, Aubert, and
  Barlaud]{charbonnier1994}
P. Charbonnier, L. Blanc-Feraud, G. Aubert, and M. Barlaud.
\newblock Two deterministic half-quadratic regularization algorithms for
  computed imaging.
\newblock In \emph{Proceedings of 1st International Conference on Image
  Processing}, pages 168--172 vol.2, 1994.

\bibitem[Dhariwal and Nichol(2021)]{dhariwal2021diffusion}
Prafulla Dhariwal and Alexander Nichol.
\newblock Diffusion models beat gans on image synthesis.
\newblock \emph{Advances in Neural Information Processing Systems},
  34:\penalty0 8780--8794, 2021.

\bibitem[Fridman et~al.(2023)Fridman, Abecasis, Kasten, and
  Dekel]{fridman2023scenescape}
Rafail Fridman, Amit Abecasis, Yoni Kasten, and Tali Dekel.
\newblock Scenescape: Text-driven consistent scene generation.
\newblock \emph{arXiv preprint arXiv:2302.01133}, 2023.

\bibitem[Gu et~al.(2023)Gu, Trevithick, Lin, Susskind, Theobalt, Liu, and
  Ramamoorthi]{gu2023nerfdiff}
Jiatao Gu, Alex Trevithick, Kai-En Lin, Josh Susskind, Christian Theobalt,
  Lingjie Liu, and Ravi Ramamoorthi.
\newblock Nerfdiff: Single-image view synthesis with nerf-guided distillation
  from 3d-aware diffusion.
\newblock \emph{arXiv preprint arXiv:2302.10109}, 2023.

\bibitem[Haque et~al.(2023)Haque, Tancik, Efros, Holynski, and
  Kanazawa]{instructnerf2023}
Ayaan Haque, Matthew Tancik, Alexei Efros, Aleksander Holynski, and Angjoo
  Kanazawa.
\newblock Instruct-nerf2nerf: Editing 3d scenes with instructions.
\newblock In \emph{Proceedings of the IEEE/CVF International Conference on
  Computer Vision}, 2023.

\bibitem[Ho and Salimans(2022)]{ho2022classifier}
Jonathan Ho and Tim Salimans.
\newblock Classifier-free diffusion guidance.
\newblock \emph{arXiv preprint arXiv:2207.12598}, 2022.

\bibitem[Ho et~al.(2020)Ho, Jain, and Abbeel]{ho2020denoising}
Jonathan Ho, Ajay Jain, and Pieter Abbeel.
\newblock Denoising diffusion probabilistic models.
\newblock \emph{Advances in Neural Information Processing Systems},
  33:\penalty0 6840--6851, 2020.

\bibitem[Ho et~al.(2022{\natexlab{a}})Ho, Chan, Saharia, Whang, Gao, Gritsenko,
  Kingma, Poole, Norouzi, Fleet, et~al.]{ho2022imagen}
Jonathan Ho, William Chan, Chitwan Saharia, Jay Whang, Ruiqi Gao, Alexey
  Gritsenko, Diederik~P Kingma, Ben Poole, Mohammad Norouzi, David~J Fleet,
  et~al.
\newblock Imagen video: High definition video generation with diffusion models.
\newblock \emph{arXiv preprint arXiv:2210.02303}, 2022{\natexlab{a}}.

\bibitem[Ho et~al.(2022{\natexlab{b}})Ho, Saharia, Chan, Fleet, Norouzi, and
  Salimans]{ho2022cascaded}
Jonathan Ho, Chitwan Saharia, William Chan, David~J Fleet, Mohammad Norouzi,
  and Tim Salimans.
\newblock Cascaded diffusion models for high fidelity image generation.
\newblock \emph{J. Mach. Learn. Res.}, 23:\penalty0 47--1, 2022{\natexlab{b}}.

\bibitem[Ho et~al.(2022{\natexlab{c}})Ho, Salimans, Gritsenko, Chan, Norouzi,
  and Fleet]{ho2022video}
Jonathan Ho, Tim Salimans, Alexey Gritsenko, William Chan, Mohammad Norouzi,
  and David~J Fleet.
\newblock Video diffusion models.
\newblock \emph{arXiv preprint arXiv:2204.03458}, 2022{\natexlab{c}}.

\bibitem[Jain et~al.(2022)Jain, Mildenhall, Barron, Abbeel, and
  Poole]{jain2022zero}
Ajay Jain, Ben Mildenhall, Jonathan~T Barron, Pieter Abbeel, and Ben Poole.
\newblock Zero-shot text-guided object generation with dream fields.
\newblock In \emph{Proceedings of the IEEE/CVF Conference on Computer Vision
  and Pattern Recognition}, pages 867--876, 2022.

\bibitem[Kasahara et~al.(2023)Kasahara, Agrawal, Engin, Chavan-Dafle, Song, and
  Isler]{kasahara2023ric}
Isaac Kasahara, Shubham Agrawal, Selim Engin, Nikhil Chavan-Dafle, Shuran Song,
  and Volkan Isler.
\newblock Ric: Rotate-inpaint-complete for generalizable scene reconstruction,
  2023.

\bibitem[Katzir et~al.(2023)Katzir, Patashnik, Cohen-Or, and
  Lischinski]{katzir2023noise}
Oren Katzir, Or Patashnik, Daniel Cohen-Or, and Dani Lischinski.
\newblock Noise-free score distillation.
\newblock \emph{arXiv preprint arXiv:2310.17590}, 2023.

\bibitem[Kingma et~al.(2021)Kingma, Salimans, Poole, and
  Ho]{kingma2021variational}
Diederik Kingma, Tim Salimans, Ben Poole, and Jonathan Ho.
\newblock Variational diffusion models.
\newblock \emph{Advances in neural information processing systems},
  34:\penalty0 21696--21707, 2021.

\bibitem[Kingma and Ba(2017)]{kingma2017adam}
Diederik~P. Kingma and Jimmy Ba.
\newblock Adam: A method for stochastic optimization, 2017.

\bibitem[Kirillov et~al.(2023)Kirillov, Mintun, Ravi, Mao, Rolland, Gustafson,
  Xiao, Whitehead, Berg, Lo, et~al.]{kirillov2023segment}
Alexander Kirillov, Eric Mintun, Nikhila Ravi, Hanzi Mao, Chloe Rolland, Laura
  Gustafson, Tete Xiao, Spencer Whitehead, Alexander~C Berg, Wan-Yen Lo, et~al.
\newblock Segment anything.
\newblock \emph{arXiv preprint arXiv:2304.02643}, 2023.

\bibitem[Le~Pendu et~al.(2018)Le~Pendu, Jiang, and Guillemot]{le2018light}
Mikael Le~Pendu, Xiaoran Jiang, and Christine Guillemot.
\newblock Light field inpainting propagation via low rank matrix completion.
\newblock \emph{IEEE Transactions on Image Processing}, 27\penalty0
  (4):\penalty0 1981--1993, 2018.

\bibitem[Lin et~al.(2023)Lin, Gao, Tang, Takikawa, Zeng, Huang, Kreis, Fidler,
  Liu, and Lin]{Lin_2023_CVPR}
Chen-Hsuan Lin, Jun Gao, Luming Tang, Towaki Takikawa, Xiaohui Zeng, Xun Huang,
  Karsten Kreis, Sanja Fidler, Ming-Yu Liu, and Tsung-Yi Lin.
\newblock Magic3d: High-resolution text-to-3d content creation.
\newblock In \emph{Proceedings of the IEEE/CVF Conference on Computer Vision
  and Pattern Recognition (CVPR)}, pages 300--309, 2023.

\bibitem[Liu et~al.(2022)Liu, Shen, and Chen]{liu2022nerfin}
Hao-Kang Liu, I-Chao Shen, and Bing-Yu Chen.
\newblock Nerf-in: Free-form nerf inpainting with rgb-d priors, 2022.

\bibitem[Liu et~al.(2023)Liu, Wu, Van~Hoorick, Tokmakov, Zakharov, and
  Vondrick]{liu2023zero}
Ruoshi Liu, Rundi Wu, Basile Van~Hoorick, Pavel Tokmakov, Sergey Zakharov, and
  Carl Vondrick.
\newblock Zero-1-to-3: Zero-shot one image to 3d object.
\newblock In \emph{Proceedings of the IEEE/CVF International Conference on
  Computer Vision}, pages 9298--9309, 2023.

\bibitem[Martin-Brualla et~al.(2021)Martin-Brualla, Radwan, Sajjadi, Barron,
  Dosovitskiy, and Duckworth]{martinbrualla2020nerfw}
Ricardo Martin-Brualla, Noha Radwan, Mehdi S.~M. Sajjadi, Jonathan~T. Barron,
  Alexey Dosovitskiy, and Daniel Duckworth.
\newblock {NeRF in the Wild: Neural Radiance Fields for Unconstrained Photo
  Collections}.
\newblock In \emph{CVPR}, 2021.

\bibitem[Meng et~al.(2023)Meng, Rombach, Gao, Kingma, Ermon, Ho, and
  Salimans]{meng2023distillation}
Chenlin Meng, Robin Rombach, Ruiqi Gao, Diederik Kingma, Stefano Ermon,
  Jonathan Ho, and Tim Salimans.
\newblock On distillation of guided diffusion models.
\newblock In \emph{Proceedings of the IEEE/CVF Conference on Computer Vision
  and Pattern Recognition}, pages 14297--14306, 2023.

\bibitem[Metzer et~al.(2023)Metzer, Richardson, Patashnik, Giryes, and
  Cohen-Or]{Metzer_2023_CVPR}
Gal Metzer, Elad Richardson, Or Patashnik, Raja Giryes, and Daniel Cohen-Or.
\newblock Latent-nerf for shape-guided generation of 3d shapes and textures.
\newblock In \emph{Proceedings of the IEEE/CVF Conference on Computer Vision
  and Pattern Recognition (CVPR)}, pages 12663--12673, 2023.

\bibitem[Mildenhall et~al.(2021)Mildenhall, Srinivasan, Tancik, Barron,
  Ramamoorthi, and Ng]{mildenhall2021nerf}
Ben Mildenhall, Pratul~P Srinivasan, Matthew Tancik, Jonathan~T Barron, Ravi
  Ramamoorthi, and Ren Ng.
\newblock Nerf: Representing scenes as neural radiance fields for view
  synthesis.
\newblock \emph{Communications of the ACM}, 65\penalty0 (1):\penalty0 99--106,
  2021.

\bibitem[Mirzaei et~al.(2023{\natexlab{a}})Mirzaei, Aumentado-Armstrong,
  Brubaker, Kelly, Levinshtein, Derpanis, and
  Gilitschenski]{mirzaei2023reference}
Ashkan Mirzaei, Tristan Aumentado-Armstrong, Marcus~A. Brubaker, Jonathan
  Kelly, Alex Levinshtein, Konstantinos~G. Derpanis, and Igor Gilitschenski.
\newblock Reference-guided controllable inpainting of neural radiance fields.
\newblock In \emph{ICCV}, 2023{\natexlab{a}}.

\bibitem[Mirzaei et~al.(2023{\natexlab{b}})Mirzaei, Aumentado-Armstrong,
  Derpanis, Kelly, Brubaker, Gilitschenski, and Levinshtein]{mirzaei2023spin}
Ashkan Mirzaei, Tristan Aumentado-Armstrong, Konstantinos~G Derpanis, Jonathan
  Kelly, Marcus~A Brubaker, Igor Gilitschenski, and Alex Levinshtein.
\newblock Spin-nerf: Multiview segmentation and perceptual inpainting with
  neural radiance fields.
\newblock In \emph{Proceedings of the IEEE/CVF Conference on Computer Vision
  and Pattern Recognition}, pages 20669--20679, 2023{\natexlab{b}}.

\bibitem[M{\"u}ller et~al.(2023)M{\"u}ller, Siddiqui, Porzi, Bulo,
  Kontschieder, and Nie{\ss}ner]{muller2023diffrf}
Norman M{\"u}ller, Yawar Siddiqui, Lorenzo Porzi, Samuel~Rota Bulo, Peter
  Kontschieder, and Matthias Nie{\ss}ner.
\newblock Diffrf: Rendering-guided 3d radiance field diffusion.
\newblock In \emph{Proceedings of the IEEE/CVF Conference on Computer Vision
  and Pattern Recognition}, pages 4328--4338, 2023.

\bibitem[M{\"u}ller et~al.(2022)M{\"u}ller, Evans, Schied, and
  Keller]{muller2022instant}
Thomas M{\"u}ller, Alex Evans, Christoph Schied, and Alexander Keller.
\newblock Instant neural graphics primitives with a multiresolution hash
  encoding.
\newblock \emph{arXiv preprint arXiv:2201.05989}, 2022.

\bibitem[Niemeyer et~al.(2022)Niemeyer, Barron, Mildenhall, Sajjadi, Geiger,
  and Radwan]{niemeyer2022regnerf}
Michael Niemeyer, Jonathan~T Barron, Ben Mildenhall, Mehdi~SM Sajjadi, Andreas
  Geiger, and Noha Radwan.
\newblock Regnerf: Regularizing neural radiance fields for view synthesis from
  sparse inputs.
\newblock In \emph{Proceedings of the IEEE/CVF Conference on Computer Vision
  and Pattern Recognition}, pages 5480--5490, 2022.

\bibitem[Park et~al.(2019)Park, Liu, Wang, and Zhu]{park2019semantic}
Taesung Park, Ming-Yu Liu, Ting-Chun Wang, and Jun-Yan Zhu.
\newblock Semantic image synthesis with spatially-adaptive normalization.
\newblock In \emph{Proceedings of the IEEE/CVF conference on computer vision
  and pattern recognition}, pages 2337--2346, 2019.

\bibitem[Philip and Drettakis(2018)]{philip2018plane}
Julien Philip and George Drettakis.
\newblock Plane-based multi-view inpainting for image-based rendering in large
  scenes.
\newblock In \emph{Proceedings of the ACM SIGGRAPH Symposium on Interactive 3D
  Graphics and Games}, pages 1--11, 2018.

\bibitem[Poole et~al.(2022)Poole, Jain, Barron, and
  Mildenhall]{poole2022dreamfusion}
Ben Poole, Ajay Jain, Jonathan~T Barron, and Ben Mildenhall.
\newblock Dreamfusion: Text-to-3d using 2d diffusion.
\newblock \emph{arXiv preprint arXiv:2209.14988}, 2022.

\bibitem[Rombach et~al.(2022)Rombach, Blattmann, Lorenz, Esser, and
  Ommer]{rombach2022high}
Robin Rombach, Andreas Blattmann, Dominik Lorenz, Patrick Esser, and Bj{\"o}rn
  Ommer.
\newblock High-resolution image synthesis with latent diffusion models.
\newblock In \emph{Proceedings of the IEEE/CVF Conference on Computer Vision
  and Pattern Recognition}, pages 10684--10695, 2022.

\bibitem[Saharia et~al.(2022{\natexlab{a}})Saharia, Chan, Chang, Lee, Ho,
  Salimans, Fleet, and Norouzi]{saharia2022palette}
Chitwan Saharia, William Chan, Huiwen Chang, Chris Lee, Jonathan Ho, Tim
  Salimans, David Fleet, and Mohammad Norouzi.
\newblock Palette: Image-to-image diffusion models.
\newblock In \emph{ACM SIGGRAPH 2022 Conference Proceedings}, pages 1--10,
  2022{\natexlab{a}}.

\bibitem[Saharia et~al.(2022{\natexlab{b}})Saharia, Chan, Saxena, Li, Whang,
  Denton, Ghasemipour, Gontijo~Lopes, Karagol~Ayan, Salimans,
  et~al.]{saharia2022photorealistic}
Chitwan Saharia, William Chan, Saurabh Saxena, Lala Li, Jay Whang, Emily~L
  Denton, Kamyar Ghasemipour, Raphael Gontijo~Lopes, Burcu Karagol~Ayan, Tim
  Salimans, et~al.
\newblock Photorealistic text-to-image diffusion models with deep language
  understanding.
\newblock \emph{Advances in Neural Information Processing Systems},
  35:\penalty0 36479--36494, 2022{\natexlab{b}}.

\bibitem[Saharia et~al.(2022{\natexlab{c}})Saharia, Ho, Chan, Salimans, Fleet,
  and Norouzi]{saharia2022image}
Chitwan Saharia, Jonathan Ho, William Chan, Tim Salimans, David~J Fleet, and
  Mohammad Norouzi.
\newblock Image super-resolution via iterative refinement.
\newblock \emph{IEEE Transactions on Pattern Analysis and Machine
  Intelligence}, 2022{\natexlab{c}}.

\bibitem[Shue et~al.(2023)Shue, Chan, Po, Ankner, Wu, and
  Wetzstein]{shue20233d}
J~Ryan Shue, Eric~Ryan Chan, Ryan Po, Zachary Ankner, Jiajun Wu, and Gordon
  Wetzstein.
\newblock 3d neural field generation using triplane diffusion.
\newblock In \emph{Proceedings of the IEEE/CVF Conference on Computer Vision
  and Pattern Recognition}, pages 20875--20886, 2023.

\bibitem[Sohl-Dickstein et~al.(2015)Sohl-Dickstein, Weiss, Maheswaranathan, and
  Ganguli]{sohl2015deep}
Jascha Sohl-Dickstein, Eric Weiss, Niru Maheswaranathan, and Surya Ganguli.
\newblock Deep unsupervised learning using nonequilibrium thermodynamics.
\newblock In \emph{International Conference on Machine Learning}, pages
  2256--2265. PMLR, 2015.

\bibitem[Song et~al.(2020)Song, Sohl-Dickstein, Kingma, Kumar, Ermon, and
  Poole]{song2020score}
Yang Song, Jascha Sohl-Dickstein, Diederik~P Kingma, Abhishek Kumar, Stefano
  Ermon, and Ben Poole.
\newblock Score-based generative modeling through stochastic differential
  equations.
\newblock \emph{arXiv preprint arXiv:2011.13456}, 2020.

\bibitem[Thonat et~al.(2016)Thonat, Shechtman, Paris, and
  Drettakis]{thonat2016multi}
Theo Thonat, Eli Shechtman, Sylvain Paris, and George Drettakis.
\newblock Multi-view inpainting for image-based scene editing and rendering.
\newblock In \emph{International Conference on 3D Vision (3DV)}, pages
  351--359. IEEE, 2016.

\bibitem[Tseng et~al.(2023)Tseng, Li, Kim, Alsisan, Huang, and
  Kopf]{tseng2023consistent}
Hung-Yu Tseng, Qinbo Li, Changil Kim, Suhib Alsisan, Jia-Bin Huang, and
  Johannes Kopf.
\newblock Consistent view synthesis with pose-guided diffusion models.
\newblock In \emph{Proceedings of the IEEE/CVF Conference on Computer Vision
  and Pattern Recognition}, pages 16773--16783, 2023.

\bibitem[Vahdat et~al.(2022)Vahdat, Williams, Gojcic, Litany, Fidler, Kreis,
  et~al.]{vahdat2022lion}
Arash Vahdat, Francis Williams, Zan Gojcic, Or Litany, Sanja Fidler, Karsten
  Kreis, et~al.
\newblock Lion: Latent point diffusion models for 3d shape generation.
\newblock \emph{Advances in Neural Information Processing Systems},
  35:\penalty0 10021--10039, 2022.

\bibitem[Wang et~al.(2023{\natexlab{a}})Wang, Zhang, Abboud, and
  Süsstrunk]{wang2023inpaintnerf360}
Dongqing Wang, Tong Zhang, Alaa Abboud, and Sabine Süsstrunk.
\newblock Inpaintnerf360: Text-guided 3d inpainting on unbounded neural
  radiance fields, 2023{\natexlab{a}}.

\bibitem[Wang et~al.(2023{\natexlab{b}})Wang, Du, Li, Yeh, and
  Shakhnarovich]{wang2023score}
Haochen Wang, Xiaodan Du, Jiahao Li, Raymond~A Yeh, and Greg Shakhnarovich.
\newblock Score jacobian chaining: Lifting pretrained 2d diffusion models for
  3d generation.
\newblock In \emph{Proceedings of the IEEE/CVF Conference on Computer Vision
  and Pattern Recognition}, pages 12619--12629, 2023{\natexlab{b}}.

\bibitem[Wang et~al.(2023{\natexlab{c}})Wang, Zhang, Zhang, Gu, Bao,
  Baltrusaitis, Shen, Chen, Wen, Chen, et~al.]{wang2023rodin}
Tengfei Wang, Bo Zhang, Ting Zhang, Shuyang Gu, Jianmin Bao, Tadas
  Baltrusaitis, Jingjing Shen, Dong Chen, Fang Wen, Qifeng Chen, et~al.
\newblock Rodin: A generative model for sculpting 3d digital avatars using
  diffusion.
\newblock In \emph{Proceedings of the IEEE/CVF Conference on Computer Vision
  and Pattern Recognition}, pages 4563--4573, 2023{\natexlab{c}}.

\bibitem[Wang et~al.(2023{\natexlab{d}})Wang, Lu, Wang, Bao, Li, Su, and
  Zhu]{wang2023prolificdreamer}
Zhengyi Wang, Cheng Lu, Yikai Wang, Fan Bao, Chongxuan Li, Hang Su, and Jun
  Zhu.
\newblock Prolificdreamer: High-fidelity and diverse text-to-3d generation with
  variational score distillation.
\newblock \emph{arXiv preprint arXiv:2305.16213}, 2023{\natexlab{d}}.

\bibitem[Warburg et~al.(2023)Warburg, Weber, Tancik, Holynski, and
  Kanazawa]{warburg2023nerfbusters}
Frederik Warburg, Ethan Weber, Matthew Tancik, Aleksander Holynski, and Angjoo
  Kanazawa.
\newblock Nerfbusters: Removing ghostly artifacts from casually captured nerfs.
\newblock \emph{arXiv preprint arXiv:2304.10532}, 2023.

\bibitem[Watson et~al.(2022)Watson, Chan, Martin-Brualla, Ho, Tagliasacchi, and
  Norouzi]{watson2022novel}
Daniel Watson, William Chan, Ricardo Martin-Brualla, Jonathan Ho, Andrea
  Tagliasacchi, and Mohammad Norouzi.
\newblock Novel view synthesis with diffusion models.
\newblock \emph{arXiv preprint arXiv:2210.04628}, 2022.

\bibitem[Weder et~al.(2023)Weder, Garcia-Hernando, Monszpart, Pollefeys,
  Brostow, Firman, and Vicente]{weder2023removing}
Silvan Weder, Guillermo Garcia-Hernando, Aron Monszpart, Marc Pollefeys,
  Gabriel~J Brostow, Michael Firman, and Sara Vicente.
\newblock Removing objects from neural radiance fields.
\newblock In \emph{Proceedings of the IEEE/CVF Conference on Computer Vision
  and Pattern Recognition}, pages 16528--16538, 2023.

\bibitem[Wei et~al.(2023)Wei, Funkhouser, and Rusinkiewicz]{wei2023clutter}
Fangyin Wei, Thomas Funkhouser, and Szymon Rusinkiewicz.
\newblock Clutter detection and removal in 3d scenes with view-consistent
  inpainting.
\newblock In \emph{Proceedings of the IEEE/CVF International Conference on
  Computer Vision}, pages 18131--18141, 2023.

\bibitem[Xiang et~al.(2023)Xiang, Yang, Huang, and Tong]{xiang20233d}
Jianfeng Xiang, Jiaolong Yang, Binbin Huang, and Xin Tong.
\newblock 3d-aware image generation using 2d diffusion models.
\newblock \emph{arXiv preprint arXiv:2303.17905}, 2023.

\bibitem[Yin et~al.(2023)Yin, Fu, Yang, and Lin]{yin2023or}
Youtan Yin, Zhoujie Fu, Fan Yang, and Guosheng Lin.
\newblock Or-nerf: Object removing from 3d scenes guided by multiview
  segmentation with neural radiance fields.
\newblock \emph{arXiv preprint arXiv:2305.10503}, 2023.

\bibitem[Yoo et~al.(2023)Yoo, Guo, Matsuo, and Gu]{yoo2023dreamsparse}
Paul Yoo, Jiaxian Guo, Yutaka Matsuo, and Shixiang~Shane Gu.
\newblock Dreamsparse: Escaping from plato's cave with 2d diffusion model given
  sparse views.
\newblock \emph{arXiv preprint arXiv:2306.03414}, 2023.

\bibitem[Yu et~al.(2018)Yu, Lin, Yang, Shen, Lu, and Huang]{yu2018generative}
Jiahui Yu, Zhe Lin, Jimei Yang, Xiaohui Shen, Xin Lu, and Thomas~S Huang.
\newblock Generative image inpainting with contextual attention.
\newblock In \emph{Proceedings of the IEEE conference on computer vision and
  pattern recognition}, pages 5505--5514, 2018.

\bibitem[Yu et~al.(2019)Yu, Lin, Yang, Shen, Lu, and Huang]{yu2019free}
Jiahui Yu, Zhe Lin, Jimei Yang, Xiaohui Shen, Xin Lu, and Thomas~S Huang.
\newblock Free-form image inpainting with gated convolution.
\newblock In \emph{Proceedings of the IEEE/CVF international conference on
  computer vision}, pages 4471--4480, 2019.

\bibitem[Zhang et~al.(2018)Zhang, Isola, Efros, Shechtman, and
  Wang]{zhang2018unreasonable}
Richard Zhang, Phillip Isola, Alexei~A. Efros, Eli Shechtman, and Oliver Wang.
\newblock The unreasonable effectiveness of deep features as a perceptual
  metric, 2018.

\bibitem[Zhou et~al.(2018)Zhou, Tucker, Flynn, Fyffe, and
  Snavely]{zhou2018stereo}
Tinghui Zhou, Richard Tucker, John Flynn, Graham Fyffe, and Noah Snavely.
\newblock Stereo magnification: Learning view synthesis using multiplane
  images.
\newblock \emph{ACM Trans. Graph. (Proc. SIGGRAPH)}, 37, 2018.

\bibitem[Zhou and Tulsiani(2023)]{zhou2023sparsefusion}
Zhizhuo Zhou and Shubham Tulsiani.
\newblock Sparsefusion: Distilling view-conditioned diffusion for 3d
  reconstruction.
\newblock In \emph{Proceedings of the IEEE/CVF Conference on Computer Vision
  and Pattern Recognition}, pages 12588--12597, 2023.

\end{thebibliography}
}
\clearpage

\section*{Appendix}
\appendix

\section{Re10k Inpainting Diffusion Model}
\begin{figure}[ht]
\centering
\begin{subfigure}[b]{\linewidth}
\centering
    \includegraphics[width=0.24\linewidth]{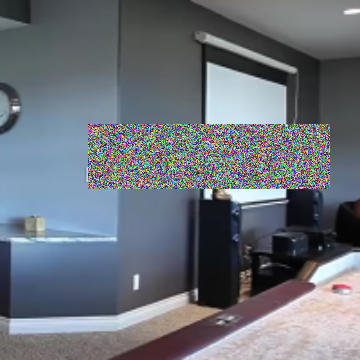}
    \includegraphics[width=0.24\linewidth]{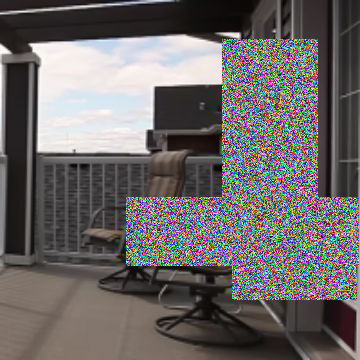}
    \includegraphics[width=0.24\linewidth]{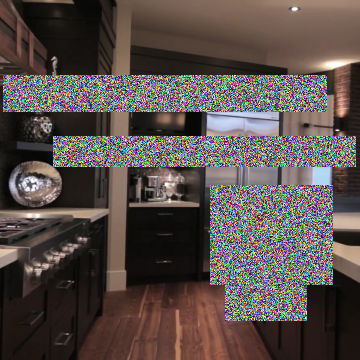}
    \includegraphics[width=0.24\linewidth]{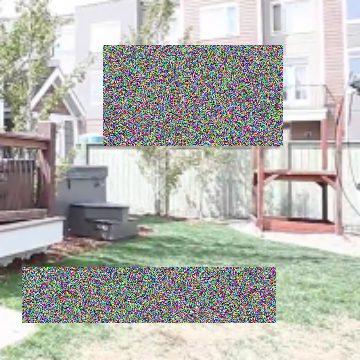}\\
    \includegraphics[width=0.24\linewidth]{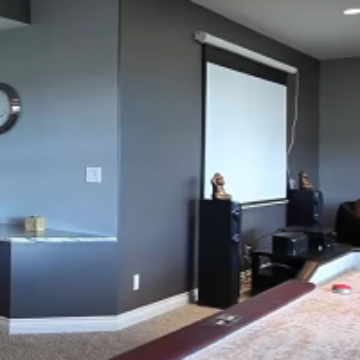}
    \includegraphics[width=0.24\linewidth]{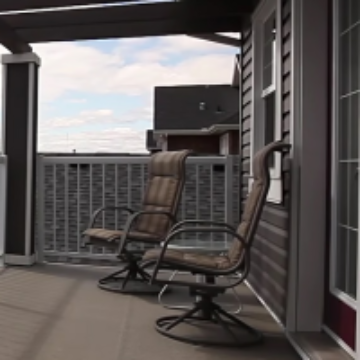}
    \includegraphics[width=0.24\linewidth]{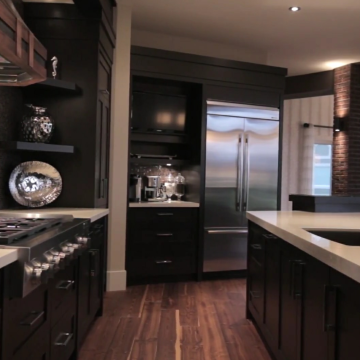}
    \includegraphics[width=0.24\linewidth]{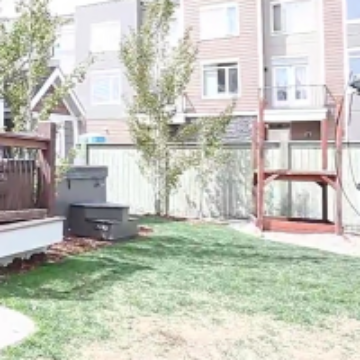}
    \caption{Inpainting Rectangles}
\end{subfigure}
\begin{subfigure}[b]{\linewidth}
\centering
    \includegraphics[width=0.24\linewidth]{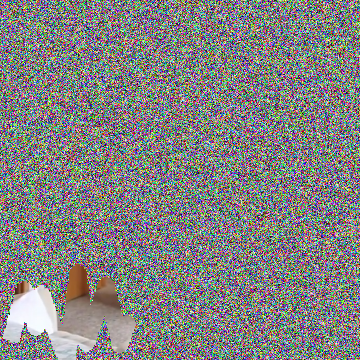}
    \includegraphics[width=0.24\linewidth]{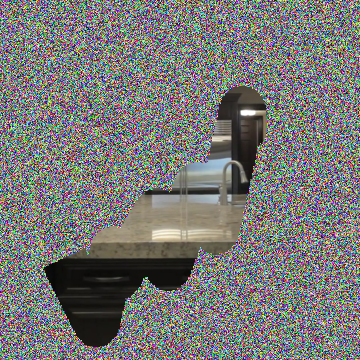}
    \includegraphics[width=0.24\linewidth]{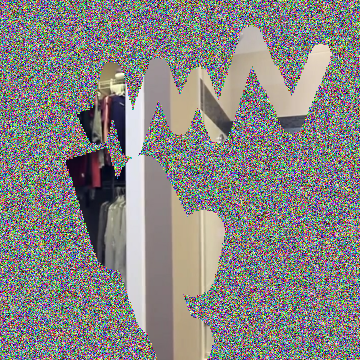}
    \includegraphics[width=0.24\linewidth]{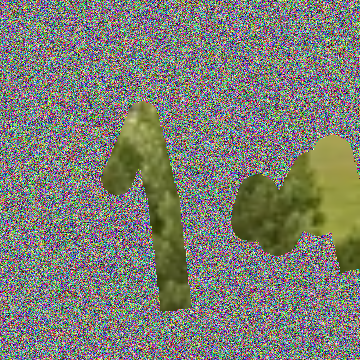} \\
    \includegraphics[width=0.24\linewidth]{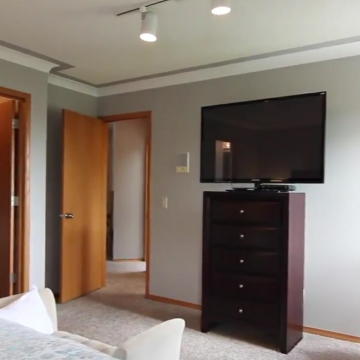}
    \includegraphics[width=0.24\linewidth]{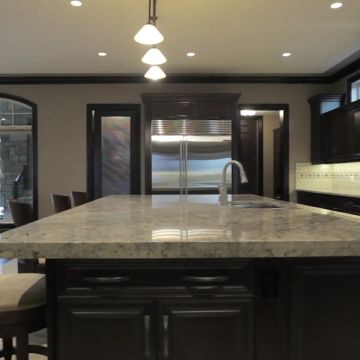}
    \includegraphics[width=0.24\linewidth]{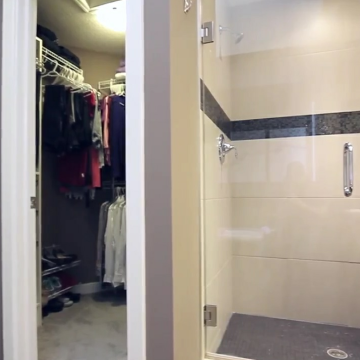}
    \includegraphics[width=0.24\linewidth]{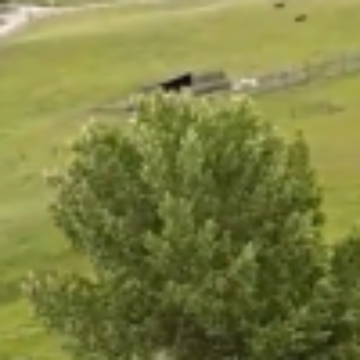}
    \caption{Outpainting Strokes}
\end{subfigure}
    \caption{The two types of masks (inpainting rectangles and outpainting strokes) and image transformations (256x256 crop or largest square crop) used to train the inpainting diffusion model. Note these are test results.}
    \label{fig:inpainting_masks}
\end{figure}

Following Palette \cite{saharia2022palette}, we train an inpainting diffusion model where the only conditioning is a masked input image. The masks denote the inpainting region and are a combination of free-form strokes which cover most of the scene (outpainting masks) and rectangles which cover a relatively smaller region of the scene (inpainting masks) as shown in Figure \ref{fig:inpainting_masks}.
Images from the Re10k dataset are resized to $1080\times 1920$ from a  native resolution of $720\times 1280$ and normalized to $[-1, 1]$. During training, each image from the dataset is preprocessed by either taking a random $256\times256$ crop or a largest square crop resized to $256\times256$ pixels.

\begin{figure}[ht]
\centering
\begin{subfigure}[b]{\linewidth}
    \includegraphics[width=0.49\linewidth]{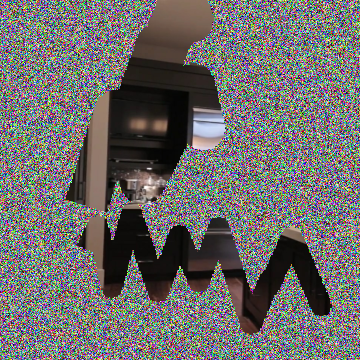}
    \includegraphics[width=0.49\linewidth]{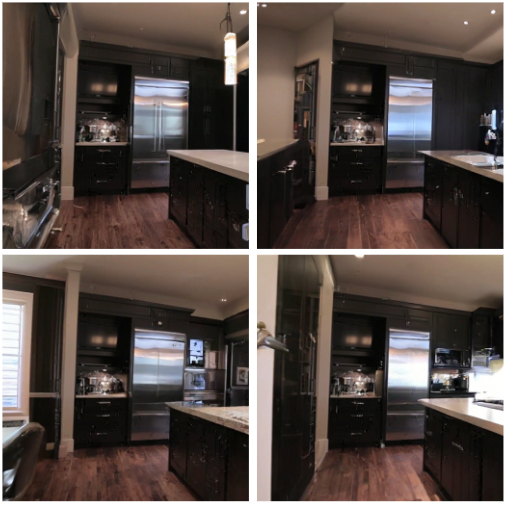}
\end{subfigure}
\begin{subfigure}[b]{\linewidth}
    \includegraphics[width=0.49\linewidth]{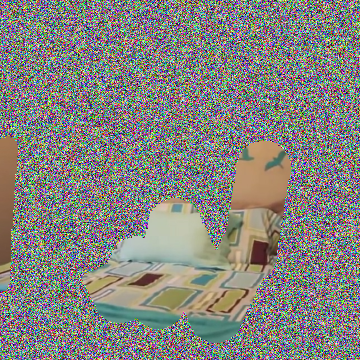}
    \includegraphics[width=0.49\linewidth]{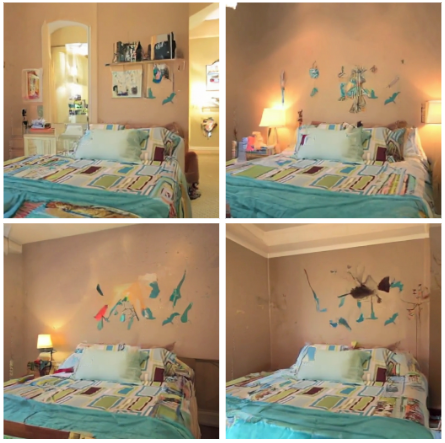}
\end{subfigure}
    \caption{Test set diffusion samples using our inpainting diffusion model. The left column shows the masked input images, and the right column shows sets of four samples from running ancestral sampling with 1000 timesteps.}
    \label{fig:diffusion_inference}
\end{figure}

We train the inpainting model for $2,250,000$ steps with a batch size of 8, on 12 Pufferfish TPUs. We use the Efficient UNet architecture described in Appendix B of Imagen \cite{saharia2022photorealistic}. During training, to implement classifier-free guidance, we drop conditioning with probability 0.1. We utilize a cosine noise schedule for training and sample continuous time steps in range $[0, 1]$. When performing ancestral sampling as in Figure \ref{fig:diffusion_inference}, we use the Denoising Diffusion Probabilistic Model (DDPM) sampler \cite{ho2020denoising} with 1000 diffusion steps. 

\section{NeRF Implementation Details}
We build on top of a simplified variant of Mip-NeRF 360 \cite{barron2021mip}, namely, making use of its proposal network and scene contraction but omitting spatial anti-aliasing / cone tracing. Following ZipNeRF \cite{barron2023zip}, we utilize hash grids from Instant-NGP \cite{muller2022instant} in place of large MLPs from Mip-NeRF 360 for efficiency, and tri-linearly resample voxel vertices at each pyramid resolution as outlined in Instant-NGP.

\subsection{Losses}
Below we provide additional details about NeRF losses $\mathcal{L}_{\text{dist}}$ and $\mathcal{L}_{\text{inter}}$ applied to all rays during each training step.

\noindent \textbf{Distortion loss:} The distortion loss from Mip-NeRF 360 helps to mitigate floaters by encouraging rays to consist of a compact set of intervals. We use the modified version from ZipNeRF where metric distances along the ray are curved with a power transformation, $\mathcal{P}(20000, -1)$, before computing the loss.

\noindent \textbf{Interlevel loss}: The interlevel loss is introduced in Mip-NeRF 360 to distill the proposal MLP online. It enforces consistency between histograms $(\mathbf{s, w})$ and $(\mathbf{\hat{s}, \hat{w}})$ over (normalized) sampled intervals and rendering weights along each ray of the NeRF and proposal MLPs respectively.

\subsection{Model Details}
We optimize with Adam \cite{kingma2017adam} with $\beta_1=0.9$, $\beta_2=0.99$, and $\epsilon=10^{-15}$, and use an exponential learning rate decay from $10^{-2}$
to $10^{-3}$ for 2.2k steps, with a warm-up in the first 100 iterations. The significant reduction in iteration count (compared to the 25K default in ZipNeRF) is due to the additional large patch SDS supervision (1) effectively increasing the batch size of the original $2^{16}$ randomly sampled rays, and (2) having the tendency to cause quality loss (over-saturate or blur) with long iteration counts.

We use two rounds of proposal sampling, sharing weights for the NGP and MLP between rounds, followed by the final NeRF round with distinct weights. Each NGP has 10 power-of-two grid scales from 16 to 8192, and a hash table size of $128^3$. A single channel per level is used for the proposal NGP, and 4 channels for the NeRF NGP. The weight decay scheme from ZipNeRF is used, penalizing the sum of the mean of grid/hash values, with the weight of 0.1 for each of the proposal and NeRF networks.

The RGB MLP has two components. First, a single layer decodes outputs from the density network (1 layer, 64 features) to ``diffuse" RGB values. Then, the view-dependent MLP (2 layers, 64 hidden units) takes bottleneck vectors (length 15) and encoded ray directions to produce the ``specular" RGB values. The two parts are added to produce the final RGB color.

\section{Inpaint-then-NeRF Experiment Details}
We pre-compute input view masks the same way as in the sphere mask experiments. To maximize similarity in the inpainted region across views, the same random noise vector is used to fill the masked region in the conditioning image, and the same random seed is reused to create the Gaussian noise input at the start of the reverse diffusion process (ancestral sampling). Neighboring images in the resulting input sequence only have small regions differing from each other, but the inconsistencies accumulate farther down the camera motion path.
During training, the LPIPS loss \cite{zhang2018unreasonable} is applied to the masked region of rendered patches from randomly sampled input views, and the reconstruction loss is applied to the randomly sampled unmasked rays following our joint optimization procedure.

\section{Additional SPIn-NeRF Results}
Figure~\ref{fig:spinnerf_supplement} shows a more comprehensive comparison of our method to SPIn-NeRF~\cite{mirzaei2023spin} using their dataset.

\section{Joint Training Hyperparameters}
We use a classifier-free guidance weight of 8 in all examples using sphere (both inpainting and outpainting) and object masks, and a guidance weight of 2 for scribble masks. We use a loss weight of 0.0001 for the patch-based depth loss, and sample four $64\times 64$ patches per GPU per step. A loss weight of 1.0 is used for the reconstruction loss and the interlevel loss, and 0.005 for distortion loss. The SDS loss is scaled according to a schedule that consists of a linear warmup which increases the loss weight from 0 to 0.1 over 100 steps, after which the loss weight is fixed at 0.1 for the rest of training. For the scribble masks, we use a reduced peak SDS loss weight of 0.04.

At the beginning of training, we apply only the NeRF reconstruction loss in the unmasked region and delay SDS optimization for the 200 steps. We choose and fix the center views for the first 50 steps of training.
We experimented with predicting diffuse and specular colors separately in the NeRF, and using only the diffuse colors for SDS. Results in Figure 1 in the main text have this enabled, while results in Figure 4 do not. 
We also experimented with early stopping at different iteration counts. Results in Figure 1 were trained for 1600 steps while results in Figure 4 were trained for 1100 steps, and outpainting scenes were trained for just 600 steps. 

\noindent \textbf{Modifications for SPIn-NeRF Dataset:}
Due to resolution, camera motion path, and nature-of-task differences we use a different set of hyperparameters for the SPIn-NeRF comparison. We use a classifier-free guidance weight of 5, SDS loss weight 2.0 with cosine decay to 1/4 of the peak value, and disable view-point annealing to in attempt to achieve an ``object removal" effect, where relatively uniform texture and smoother geometry is produced in the masked region. 
We also delay SDS to after 400 steps of NeRF training in the unmasked region, and early-stop at 1.2K iterations to avoid over saturation. We omit the patch-based depth loss for this dataset.

\section{Depth Loss Ablation}
\begin{figure}[ht]
\centering
\begin{subfigure}[b]{0.8\linewidth}
    \includegraphics[width=0.49\linewidth]{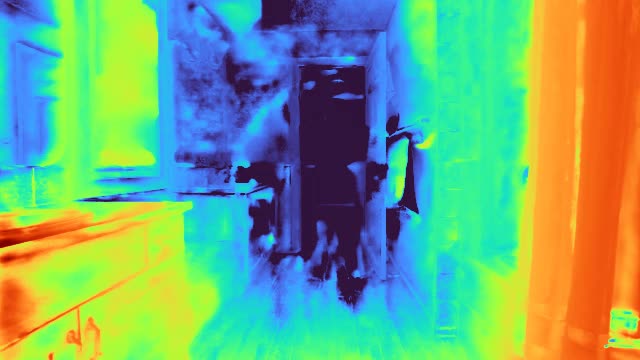}
    \includegraphics[width=0.49\linewidth]{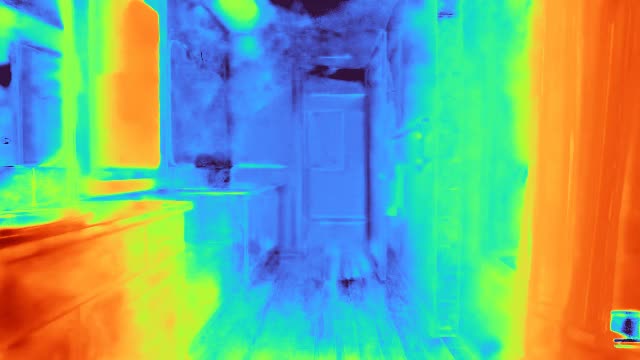}
\end{subfigure}
\begin{subfigure}[b]{0.8\linewidth}
    \includegraphics[width=0.49\linewidth]{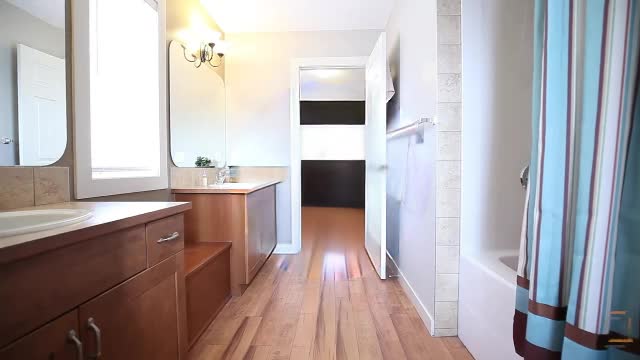}
    \includegraphics[width=0.49\linewidth]{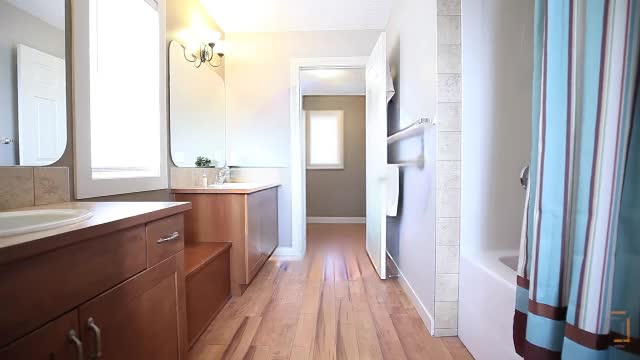}
\end{subfigure}
    \caption{Results without (left) vs with (right) depth loss. The depth loss improves 3D consistency of the inpainted results.}
    \label{fig:depth_ablation}
\end{figure}

\begin{figure*}[ht]
\centering
\includegraphics[width=\linewidth]{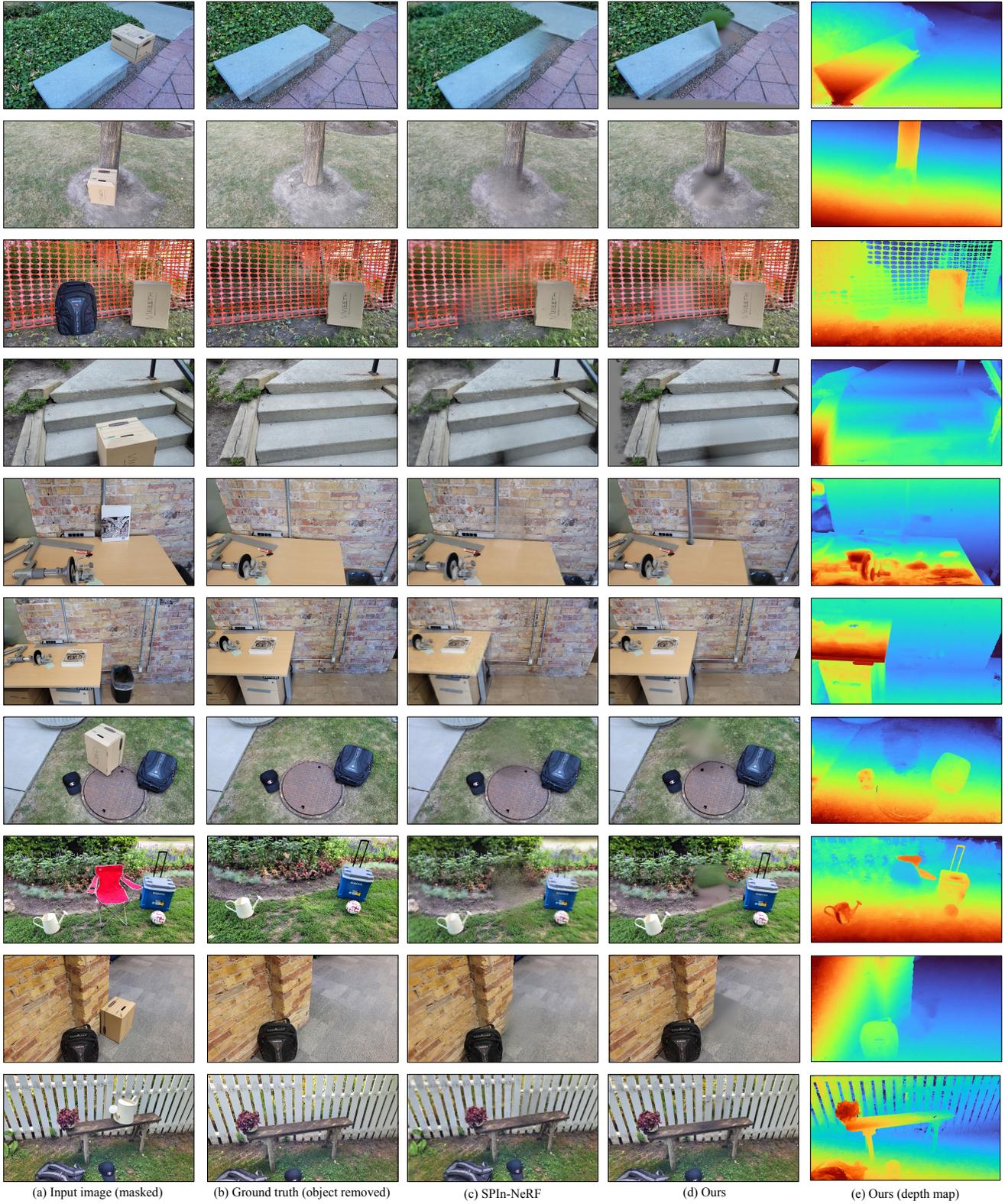}
\caption{More examples comparing our method to results from SPIn-NeRF~\cite{mirzaei2023spin}, using the SPIn-NeRF dataset.}
\label{fig:spinnerf_supplement}
\end{figure*}

We found that the use of the patch-based depth regularizer from RegNeRF  \cite{niemeyer2022regnerf} significantly improves the overall depth map quality and 3D consistency of results. A comparison of our inpainted depth maps with and without the depth loss is shown in Figure \ref{fig:depth_ablation}.

\end{document}